%% file: Sections/main.tex
\documentclass[11pt,a4paper,twocolumn]{article}
\usepackage[left=0.75in, right=0.75in, top=1in, bottom=1in]{geometry}

\usepackage{latexsym}
\usepackage{microtype}
\usepackage{fontspec}
\usepackage[round]{natbib}
\usepackage{titlesec}
\titleformat{\section}{\normalfont\large\bfseries}{\thesection}{1em}{}
\titleformat{\subsection}{\normalfont\normalsize\bfseries}{\thesubsection}{1em}{}
\titleformat{\subsubsection}{\normalfont\normalsize\bfseries}{\thesubsubsection}{1em}{}

\usepackage[table]{xcolor}
\definecolor{lightgreen}{HTML}{E8F5E9}
\definecolor{lightred}{HTML}{FFEBEE}
\definecolor{triagecolor}{RGB}{232,212,239}
\definecolor{compcolor}{RGB}{253,222,222}
\definecolor{mixcolor}{RGB}{255,243,205}
\definecolor{codebg}{HTML}{F5F5F5}

\usepackage{tikz}
\usetikzlibrary{shapes.geometric, arrows.meta, positioning, fit, backgrounds}
\usepackage{booktabs}
\usepackage{colortbl}
\usepackage{amssymb}
\usepackage{dsfont}
\usepackage{framed}

\usepackage{listings}
\lstdefinestyle{promptstyle}{
  backgroundcolor=\color{codebg},
  frame=single,
  rulecolor=\color{gray},
  basicstyle=\ttfamily\tiny,
  breaklines=true,
  breakatwhitespace=false,
  columns=fullflexible,
  keepspaces=true,
  showstringspaces=false,
  aboveskip=0.5em,
  belowskip=0.5em,
  xleftmargin=0.5em,
  xrightmargin=0.5em,
}

\usepackage{tabularx}
\usepackage{multirow}
\usepackage{makecell}
\usepackage{float}
\usepackage{caption}
\usepackage{subcaption}
\usepackage{enumitem}
\usepackage{footnote}

\usepackage{hyperref}
\hypersetup{
  colorlinks=true,
  linkcolor=blue!60!black,
  citecolor=blue!60!black,
  urlcolor=blue!60!black,
}

\usepackage[most]{tcolorbox}

\newcommand{\reasonhl}[1]{}
\tcbset{
  takeaway/.style={
    enhanced,
    colback=black!2,
    frame hidden,
    boxrule=0pt,
    arc=0pt,
    left=3pt,right=3pt,
    top=2pt,bottom=2pt,
    boxsep=1pt,
    before skip=4pt,
    after skip=4pt
  }
}

\begin{document}

\twocolumn[%
  \vspace{0.5em}\hrule\vspace{0.8em}
  \begin{center}
    {\Large\bfseries Script Gap: Evaluating LLM Triage on Indian Languages in Native vs Romanized Scripts in a Real World Setting\par}
  \end{center}
  \vspace{0.5em}\hrule\vspace{1em}
  \begin{center}
    {\normalsize
      Manurag Khullar$^*$ \quad
      Utkarsh Desai \quad
      Poorva Malviya \quad
      Aman Dalmia \quad
      Zheyuan Ryan Shi$^*$ \\[4pt]
      $^*$School of Computing and Information, University of Pittsburgh
    \par}
    \vspace{1.5em}
  \end{center}
]

 
\begin{abstract}

Large Language Models (LLMs) are increasingly deployed in high-stakes clinical applications in India. Speakers of Indian languages frequently communicate using romanized text rather than native scripts, yet existing research rarely quantifies or evaluates this orthographic variation in real world applications. We investigate how romanization impacts the reliability of LLMs in a critical domain: maternal and newborn healthcare triage. We benchmark leading LLMs on a real world dataset of user-generated health queries spanning five Indian languages and Nepali. Our results reveal consistent degradation in performance for romanized messages, with gap reaching up to 24 points across languages and models. We propose and evaluate an Uncertainty-based Selective Routing method to close this script gap. At our partner maternal health organization alone, this gap could cause nearly 2 million excess errors in triage. Our findings highlight a critical safety blind spot in LLM-based health systems: models that appear to understand romanized input may still fail to act on it reliably.

\end{abstract}
\input{Sections/intro}

\input{Sections/related}
\input{Sections/dataset}

\input{Sections/experiment}

\input{Sections/result}

\input{Sections/diagnosis}

\input{Sections/mitigation}
\input{Sections/discussion}

\input{Sections/limitations}
\bibliographystyle{plainnat}
\bibliography{references}
\clearpage
\appendix           
\input{Sections/appendix.tex}

\end{document}

%% file: Sections/intro.tex
\section{Introduction and Related Work}
Access to timely and reliable healthcare in many communities in the global south is severely constrained~\cite{mehta2024human}. 
Shortages of medical professionals, overburdened primary-care systems, and linguistic diversity limit patients’ ability to obtain high-quality guidance. Large language models (LLMs) offer a promising path to expand scalable, multilingual health communication where traditional services are stretched thin. LLMs are already being explored for patient-facing applications, such as message triage, portal-message prioritization, and safety-oriented medical question answering \citep{liu2025detecting,masanneck2024triage,borkowski2023chatgpt,singhal2025medpalm,kaur2025portal}. These systems aim to flag emergencies in patient portal messages, route complex queries to clinicians, and generate draft replies that reduce message burden for care teams \citep{ren2024primaryconcerns,chen2025raec}. 

However, existing LLMs still exhibit limitations that directly affect their suitability for clinical use. They can perform inconsistently across languages, hallucinate medical facts, and exhibit unstable reasoning under minor input perturbations. These issues are exacerbated for low-resource languages and informal, patient-generated text, where variation in spelling, phrasing, and code-mixing can substantially degrade model reliability.

A growing body of healthcare-focused LLM benchmarks aims to assess these risks, but they fall short in several ways. Many are sourced from clinical literature, or expert-curated exams, and feature synthetic question answering, rather than real patient-provider conversations~\cite{medqa,MedMCQA,jin-etal-2019-pubmedqa,arora2025healthbench}. Only a small number of datasets feature real conversations but are limited to English and Chinese~\cite{manes-etal-2024-k,he2020meddialog}, and virtually none assess triage, arguably the task most relevant for early patient engagement. 
As a result, current benchmarks fail to capture the linguistic and situational diversity encountered in frontline care.
No wonder a growing body of work shows LLMs can produce clinically unsafe or inconsistent advice, even when overall accuracy on QA benchmarks is high~\citep{diekmann2025safetyqa,draelos2025unsafeqa,bedi2024testing}.  

Furthermore, beyond the healthcare setting, handling informal, mixed-script communication at scale remains a challenge for multilingual LLMs. For Indian languages in particular, romanized input is pervasive in online communications, yet existing multilingual benchmarks minimally cover them~\cite{verma2025milu,kumar2022indicnlg,singh2024indicgenbench,singh2025indic}. Script representation is often studied in highly curated and parallel transliterated corpora, but not on noisy patient messages~\cite{husain2024romansetu,wang2024dualscript}. Critically, no existing benchmark quantifies how prevalent romanized text actually is relative to native script in real world user interactions beyond social media usage. This is a critical gap: without empirical evidence on the proportion of romanized versus native-script usage in real world LLM applications at population scale, the field lacks the basis to assess how much of the user population is underserved by native-script-focused models. Our real world corpus provides this evidence for the first time in a healthcare setting: in Hindi and Telugu, the two largest language groups in our dataset, romanized messages account for 83\% and 73\% of inputs, respectively, demonstrating that users predominantly communicate in the very script that most benchmarks overlook.

In this paper, we provide the first benchmarking evaluation of real world LLM triage performance on Indian languages across both native and romanized scripts. We make four major contributions:
(1) We construct a multilingual, multi-script healthcare triage dataset reflecting authentic patient-provider conversations.
(2) We benchmark several leading LLMs on this dataset, revealing for the first time substantial script-dependent disparities.
(3) We conduct exhaustive error analyses to uncover where and why models disproportionately fail on romanized inputs. We point out that romanized inputs lead to brittle output, even when
models seem to generate rationales that capture the user's intent.
(4) We propose an uncertainty-based selective routing method to close this script gap, and demonstrate its performance across various languages and models.

We partner with Maternal Health Organization A\footnote{Anonymized organization name for peer review.}, a nonprofit in India whose programs have reached over 41 million caregivers and patients across nine Indian states since 2014.
This study builds on real world patient-provider message data collected as part of Maternal Health Organization A’s pilot studies for evaluating LLM safety, where triage remains a critical step in early engagement.
With Maternal Health Organization A alone, the script gap revealed in this work could cause nearly 2 million excess errors in LLM-based triage.
More broadly, our work is also applicable to the many other healthcare platforms currently exploring LLM-powered solutions to enhance care delivery. We provide a concrete framework for how to evaluate triage performance and insight into the script gap in frontier models.

%% file: Sections/dataset.tex
\section{Dataset and Task}

\subsection{Clinical Setting and Corpus}
\label{subsec:clinical-setting}
\begin{figure}[t]
    \centering
    \includegraphics[width=\columnwidth]{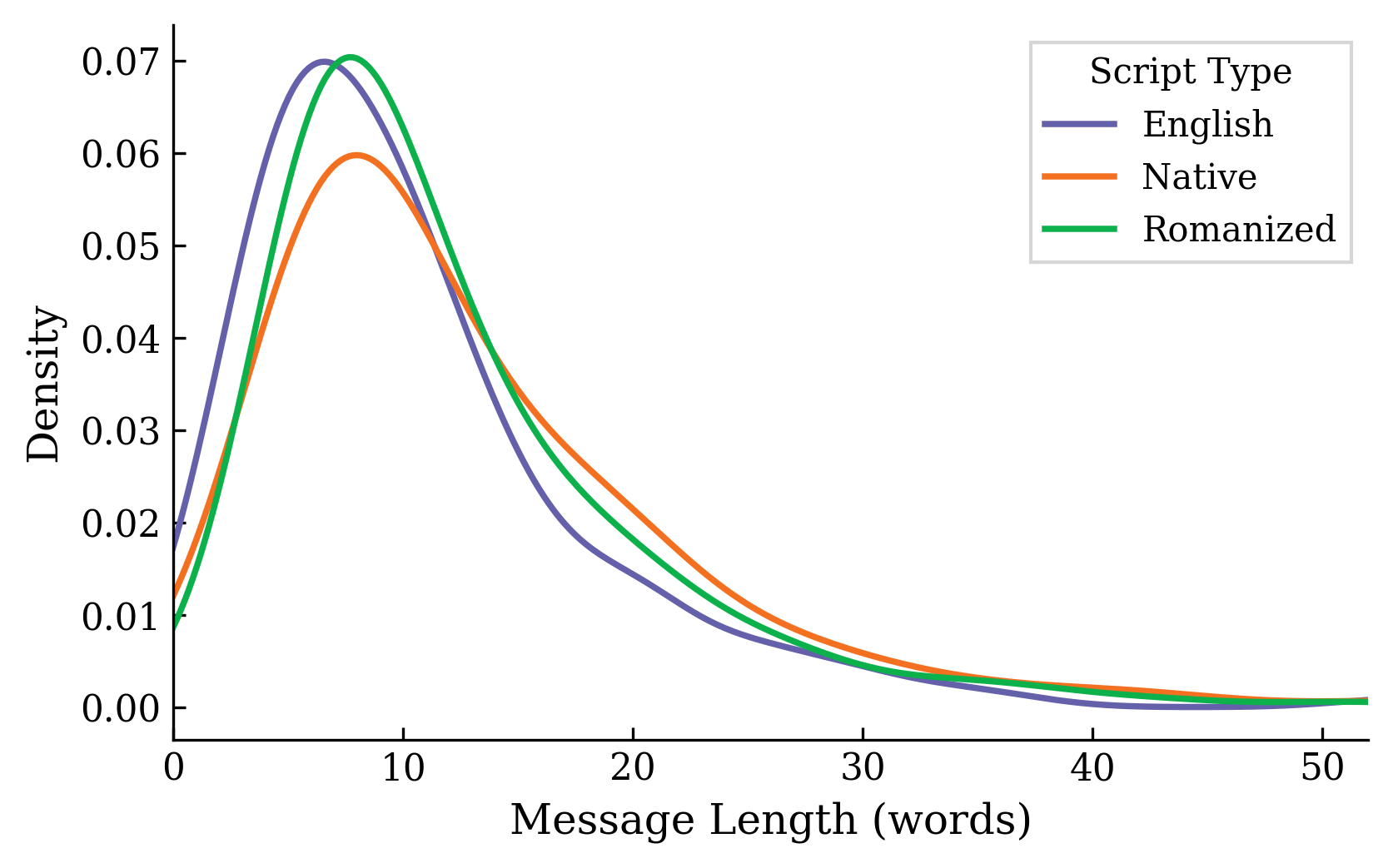}
    \caption{Kernel density estimates of message length in words across script types.}
    \label{fig:word_length_distribution}
\end{figure}

We study a de-identified corpus of short WhatsApp messages related to maternal and newborn care provided by Maternal Health Organization A. Users, typically mothers, pregnant women, or caregivers, use a WhatsApp chat interface to submit free-text queries about pregnancy and newborn issues. For this work, we restrict attention to a fixed observation window and use only de-identified data, yielding a corpus of approximately 133k conversations spanning English, five Indian languages: Hindi, Telugu, Kannada, Marathi, and Punjabi, and Nepali. From this pool, we draw a stratified random sample of 3{,}153 single-turn user messages for experiments, ensuring coverage across languages and message lengths. General statistics of the dataset are described in Table~\ref{tab:lang_native_roman_counts} and Figure~\ref{fig:word_length_distribution} .

\definecolor{nativecol}{HTML}{F37021}
\definecolor{romancol}{HTML}{0DB12D}
\newcommand{\scriptbar}[2]{%
  \begin{tikzpicture}[baseline=-0.5ex]
    \fill[nativecol] (0,0) rectangle (#1*2,0.25);
    \fill[romancol] (#1*2,0) rectangle (2,0.25);
  \end{tikzpicture}%
}

\begin{table}[t]
\centering
\small
\caption{Language distribution by script types. English serves as baseline. \textcolor{nativecol}{$\blacksquare$}\,Native \textcolor{romancol}{$\blacksquare$}\,Romanized.}
\label{tab:lang_native_roman_counts}
\setlength{\tabcolsep}{3pt}
\begin{tabular}{l r r r c}
\toprule
\textbf{Language} & \textbf{Native} & \textbf{Roman.} & \textbf{Total} & \textbf{\% N. / \% R.} \\
\midrule
English & -- & -- & 321 & -- \\
Hindi   & 158 & 702 & 860  & \scriptbar{0.184}{0.816} \\
Telugu  & 140 & 383 & 523  & \scriptbar{0.268}{0.732} \\
Kannada & 253 & 210 & 463  & \scriptbar{0.546}{0.454} \\
Marathi & 253 & 165 & 418  & \scriptbar{0.605}{0.395} \\
Punjabi & 129 & 227 & 356  & \scriptbar{0.362}{0.638} \\
Nepali  & 125 &  87 & 212  & \scriptbar{0.590}{0.410} \\
\midrule
\textbf{Total} & 1,058 & 1,774 & 3,153 & \\
\bottomrule
\end{tabular}
\end{table}

\subsection{Triage Labels and Definition}

The triage task uses three mutually exclusive labels: \emph{Emergency}, \emph{Non-emergency}, and \emph{Insufficient Information}. \emph{Emergency} covers messages that indicate, or could plausibly indicate, symptoms requiring urgent medical attention; \emph{Non-emergency} covers concerns that clearly do not require urgent care such as routine diet, breastfeeding, sleep, or administrative questions and \emph{Insufficient Information} is reserved for vague messages where it is not possible to determine emergency status such as ``I am a BP (Blood Pressure) patient". Label statistics for script types are in Figure~\ref{fig:label_distribution} and detailed definition in Appendix~\ref{app:triage-guidelines}. All messages were labeled via human annotation, cross-validated by independently labeling a stratified sample of 200 messages, yielding an overall agreement rate of 87.5\% and Cohen's $\kappa$ of 0.74. Annotation process and details by language are in Appendix~\ref{app:annotation-guidelines}.

\begin{figure}[t]
    \centering
    \includegraphics[width=\columnwidth]{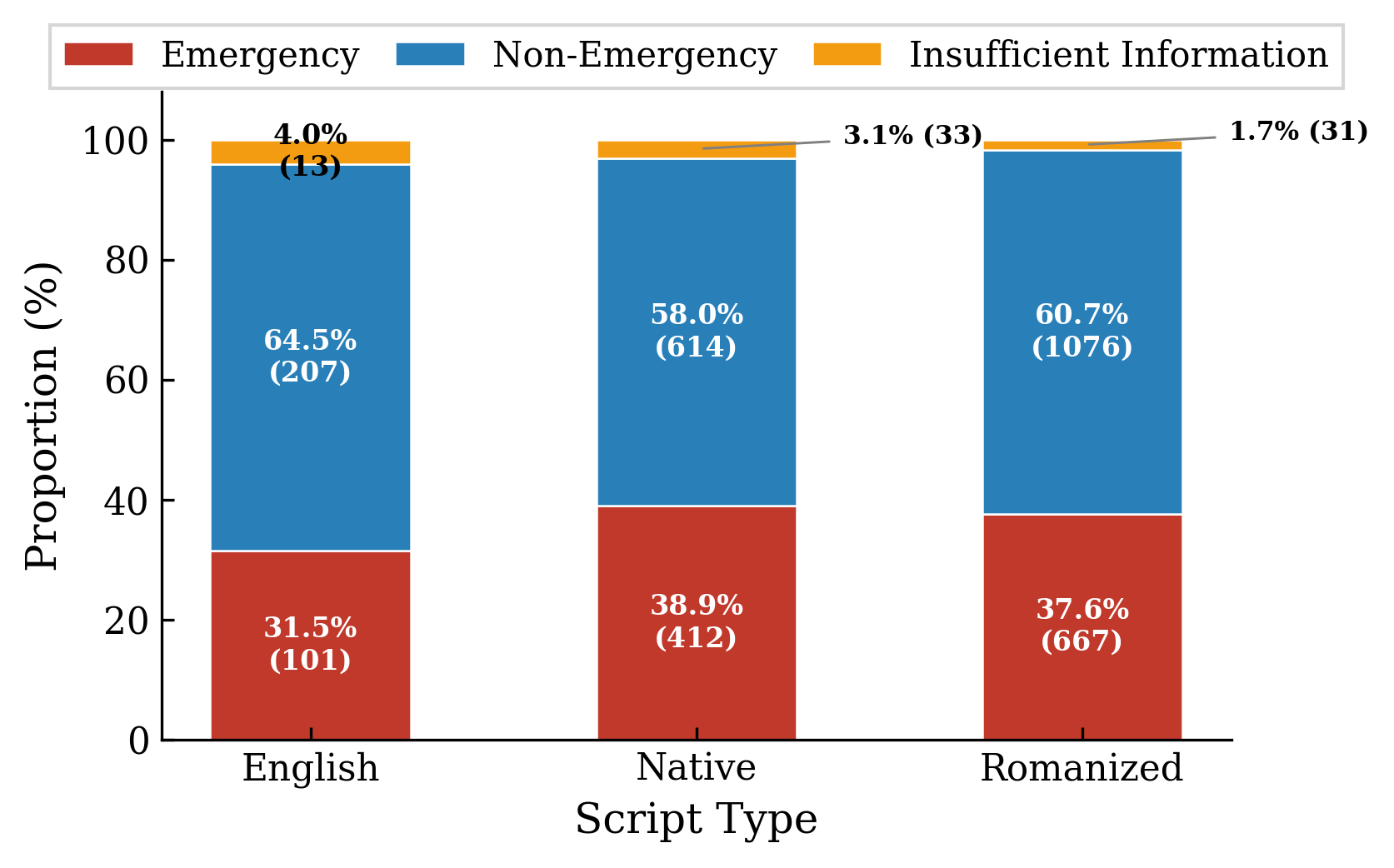}
    \caption{Distribution of human annotated labels across script types.}
    \label{fig:label_distribution}
\end{figure}

\subsection{Validation Dataset}
To assess whether the trends generalize beyond our real world clinical corpus, we construct a parallel evaluation set from HealthBench's   emergency referrals theme~\cite{arora2025healthbench}, where physicians categorized 453 prompts into three triage classes: Emergent, Conditional Emergent, and Non-Emergent. Since HealthBench is predominantly English, we use GPT-4o to translate each of the 453 prompts into native script and romanized variants for all six languages, yielding a controlled parallel corpus in which clinical content is held constant across scripts and languages. This controlled setup eliminates confounds present in our real world corpus, allowing us to isolate the effect of script choice on model performance.

%% file: Sections/experiment.tex
\section{Experimental Setup}

\subsection{Task}
The primary objective is triage classification: given a user message, the system must assign one of three triage labels. All models operate on the raw user messages, without normalization, and are evaluated using a single fixed prompt template, temperature = 0, and full context to ensure parity across models. In addition to the triage label, models are instructed to produce a brief reasoning behind the predicted label. 

\subsection{Models}

We evaluate a variety of models known for their multilingual capabilities,  organizing them into three buckets: frontier closed models include GPT-4o, Claude Sonnet 4.5 (ClaudeS4.5), large open-weight models include DeepSeek~V3, LLaMA~-4 Maverick 17B (LLaMA-4), and Qwen3-Next-80B-A3B-Instruct (Qwen3-80B), and a compact-plus-Indic-specialized bucket include Mixtral-7B, Qwen2.5-7B and Sarvam-M 24B. 




\subsection{Prompt Template}

To select the final fixed prompt, we conducted a minimal
prompt-design sweep on a held-out subset of 500 randomly
sampled messages on real world dataset, comparing four strategies:
(i) zero-shot prompt; (ii) few-shot prompt with labelled examples; (iii) Triage rules + knowledge base (KB) of emergency symptoms with chain of thought prompt. (iv) Triage rules + KB with direct prompt instructing the model to directly generate a label. In our experiment, Triage rules + KB with direct prompt consistently achieved a higher weighted F1 and emergency recall (Appendix ~\ref{appendix: promptopt}). We therefore fix the Triage Rules + KB direct template in Figure~\ref{fig:prompt-overview} for all experiments (detailed prompt in Appendix~\ref{appendix:triage-prompt-realworld}) and a similar prompt is also fixed for HealthBench dataset (Appendix~\ref{appendix:healthbench-prompt}).

\begin{figure}[t]
\centering
\small
\begin{tcolorbox}[colback=gray!5, colframe=gray!40, boxrule=0.4pt,
  left=4pt, right=4pt, top=4pt, bottom=4pt]
\textbf{Fixed Prompt Template}\\[3pt]
\textit{Role:} Multilingual triage classifier\\[1.5pt]
\textit{Label Space} Emergency $\mid$ Non-Emergency $\mid$ Insufficient Info\\[1.5pt]
\textit{Triage Rules:} Decision priorities \& edge cases\\[1.5pt]
\textit{Knowledge Base:} Emergency symptom list\\[1.5pt]
\textit{Input:} User message \\[1.5pt]
$\Rightarrow$ \textbf{Output:} Label
\end{tcolorbox}
\caption{Triage Rules + KB Direct Prompt.}
\label{fig:prompt-overview}
\end{figure}

%% file: Sections/result.tex
\begin{figure*}[t]
    \centering
    \begin{minipage}{0.64\textwidth}
        \centering
        \includegraphics[width=\textwidth]{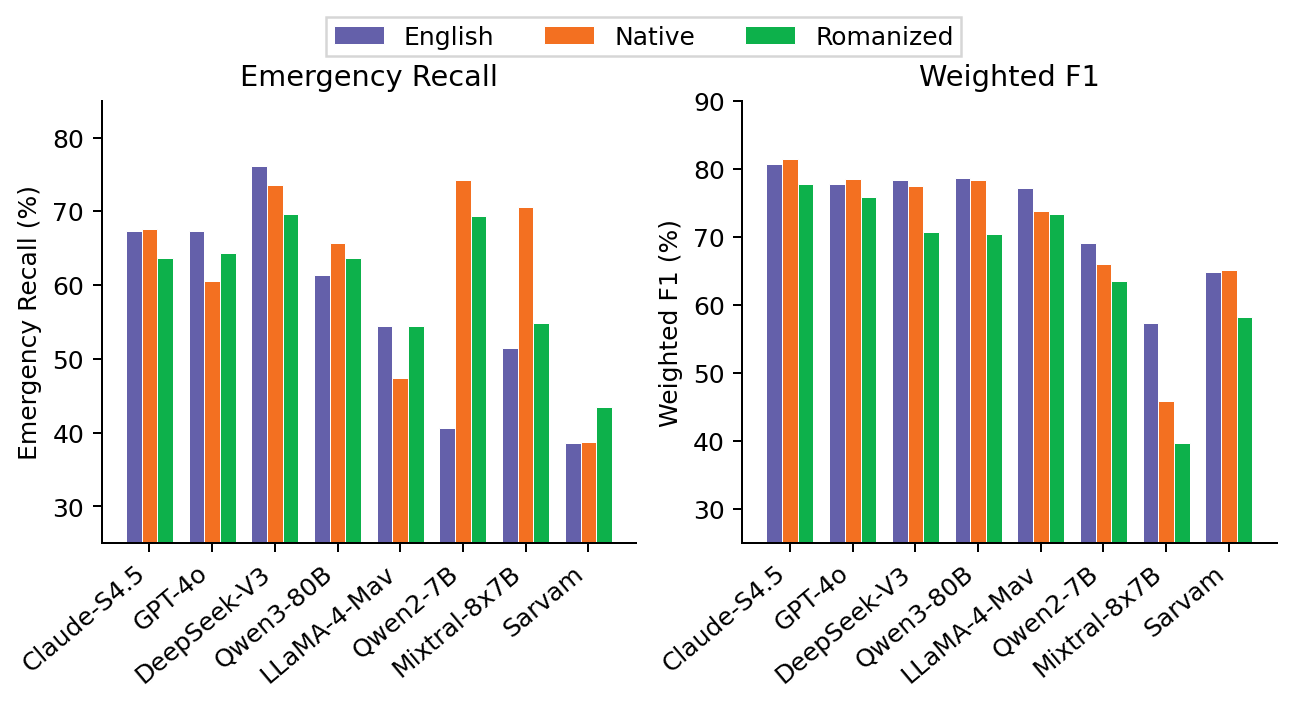}
    \end{minipage}%
    \hfill
    \begin{minipage}{0.34\textwidth}
        \centering
        \includegraphics[width=\textwidth]{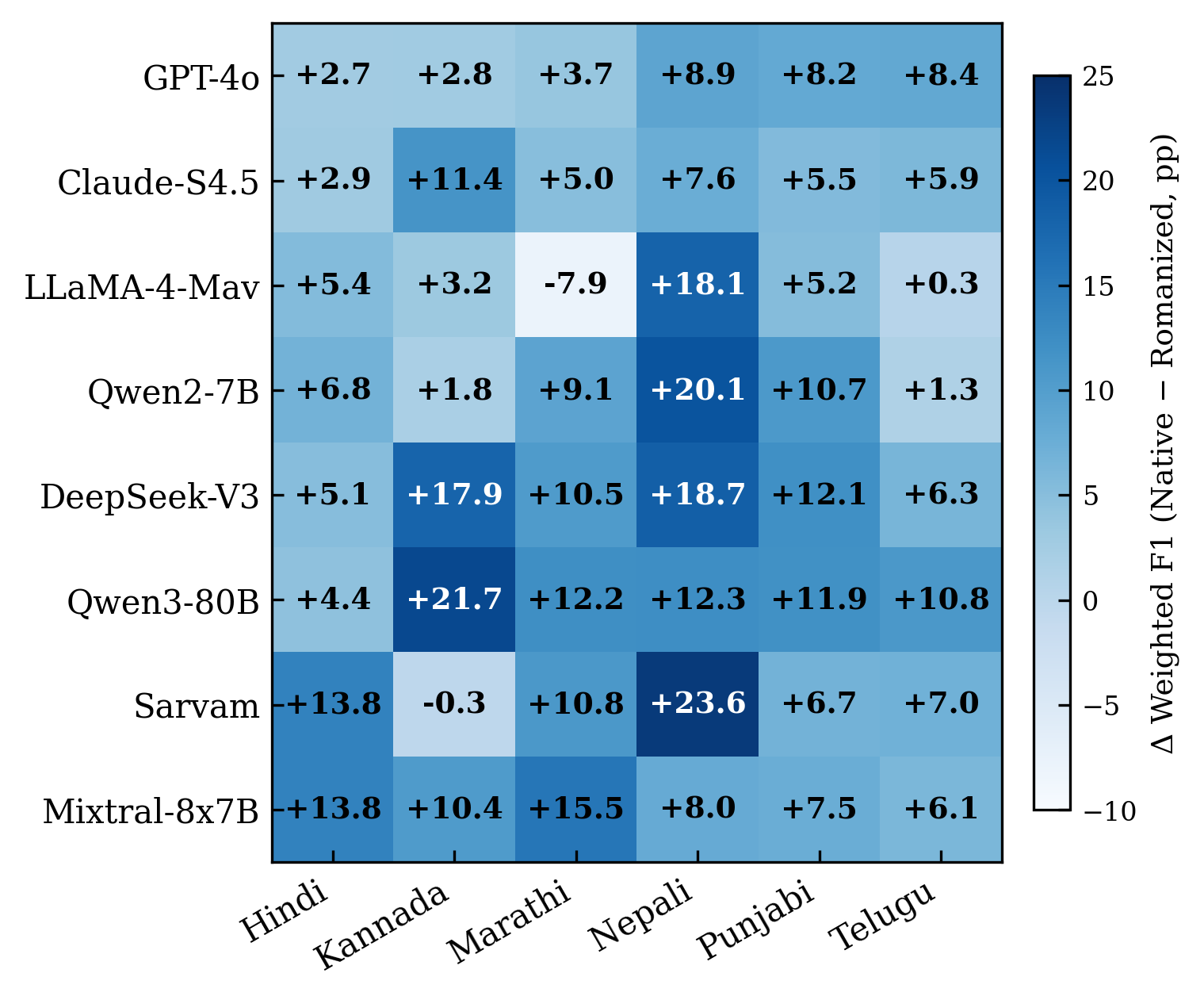}
    \end{minipage}
    \caption{\textbf{Real World Data:}(Left)~Emergency Recall and w-F1 scores across models for English, Native, and Romanized scripts. (Right)~Performance gap (Native $-$ Romanized) in w-F1 (pp) across languages and models.}
    \label{fig:benchmark_results}
\end{figure*}

\begin{figure*}[t]
    \centering
    \begin{minipage}{0.64\textwidth}
        \centering
        \includegraphics[width=\textwidth]{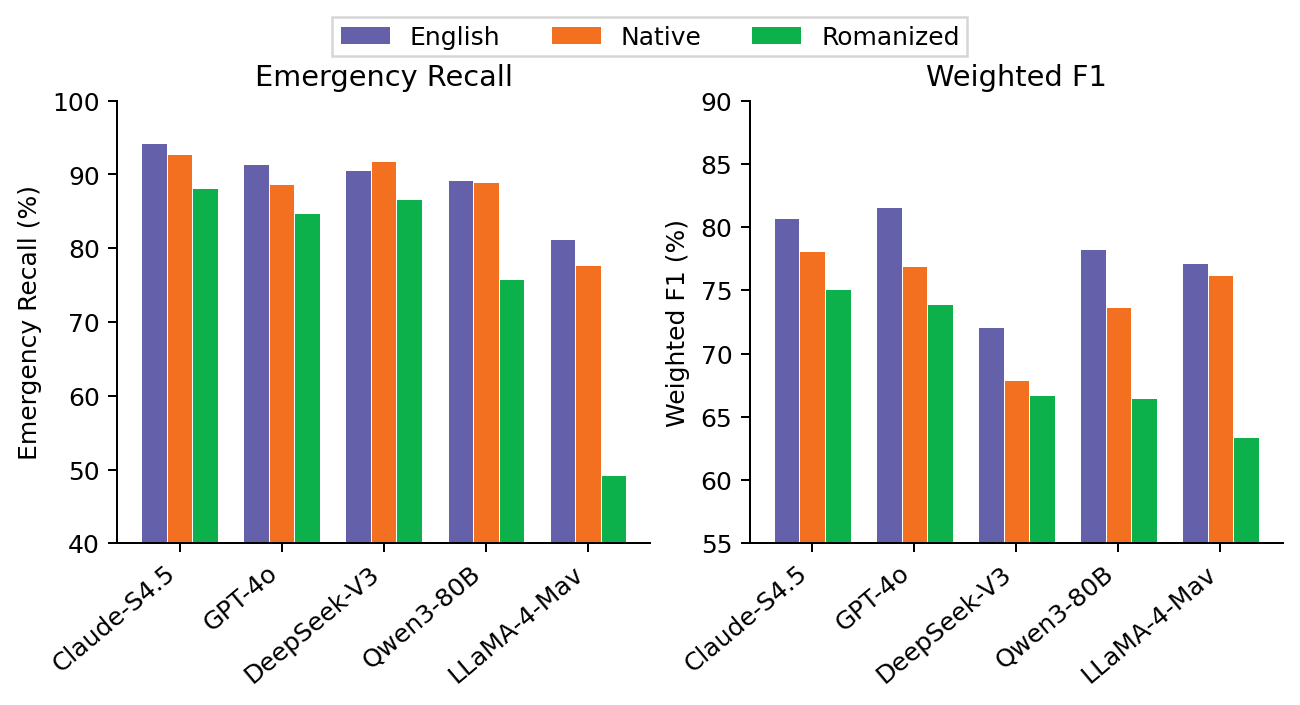}
    \end{minipage}%
    \hfill
    \begin{minipage}{0.34\textwidth}
        \centering
        \includegraphics[width=\textwidth]{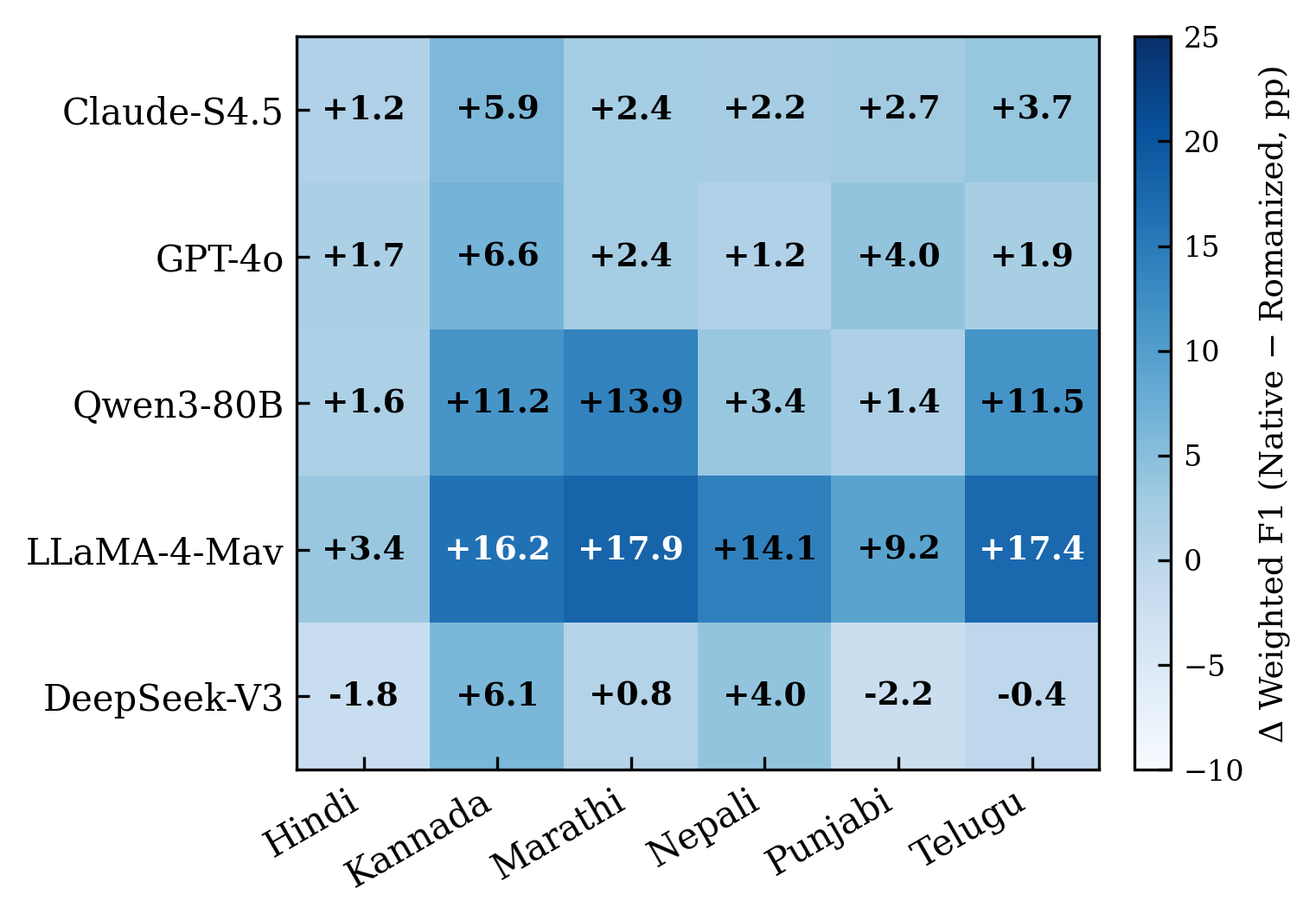}
    \end{minipage}
    \caption{\textbf{Validation on HealthBench:}~(Left)~Emergency Recall and w-F1 scores across models for English, Native, and Romanized scripts on the HealthBench parallel corpus. (Right)~Performance gap (Native $-$ Romanized) in w-F1 (pp) across languages and models.}
    \label{fig:healthbench_results}
\end{figure*}

\section{Results}
\subsection{Overall Benchmarking}
\label{sec:overall-benchmarking}

Figure~\ref{fig:benchmark_results} summarizes emergency recall and weighted F1 (w-F1) scores by model and script type. The large frontier and open-weight models---Claude-S4.5, GPT-4o, Qwen3-80B, DeepSeek-V3, and LLaMA-4---form a top tier, with w-F1 in the high 70s to low 80s. Qwen2-7B and Sarvam occupy a mid-tier roughly 10--15\,points behind, while Mixtral-8$\times$7B lags substantially further. Across w-F1 and every model, romanized input yields the weakest performance. Native-script results closely track or slightly surpass English for most models, but romanized w-F1 drops anywhere from 1 to 8\,points below native among the top-tier models on average, and this gap widens considerably for smaller models. Even Sarvam, specifically optimized for Indian languages, drops nearly 7\,points in w-F1 from native to romanized input.

\subsection{Benchmark by Language}
Figure~\ref{fig:benchmark_results} heatmap breaks this trend down by language and reveals substantial
variation. Nepali consistently shows the largest native--romanized drops across models, often exceeding 15\ points and reaching as high as 24\,points. Kannada and Telugu also exhibit pronounced degradations, with some models losing 20--30 points in accuracy when switching from native script to Roman. Marathi and Punjabi show moderate losses for most models, though certain model--language pairs diverge: Punjabi gaps are small for Claude-S4.5 and LLaMA-4 but considerably larger for Qwen3-80B and DeepSeek-V3.

\subsection{Validation on HealthBench}
\label{sec:healthbench-validation}
Figure~\ref{fig:healthbench_results} highlights that script gap persists on HealthBench. Across all five models, romanized w-F1 trails native-script w-F1, with an average drop of 5.4\,pp. HealthBench controls for clinical content, as each user message is identical across scripts. This confirms that gap observed in Section~\ref{sec:overall-benchmarking} is orthographic rather than a consequence of other confounders, such as typing noise, content, in real world data in romanized script.  Claude-S4.5 exhibit stable gaps across datasets showing approximately 3 points drop, closely mirroring real world data evaluations. Within Indian languages, higher resource languages such as Hindi and Punjabi have closing gaps in HealthBench vis-à-vis real world data, while Kannada and Telugu have large gaps on both datasets. 
\subsection{Cross Model Agreement}
\begin{figure}[t]
    \centering
    \includegraphics[width=\linewidth]{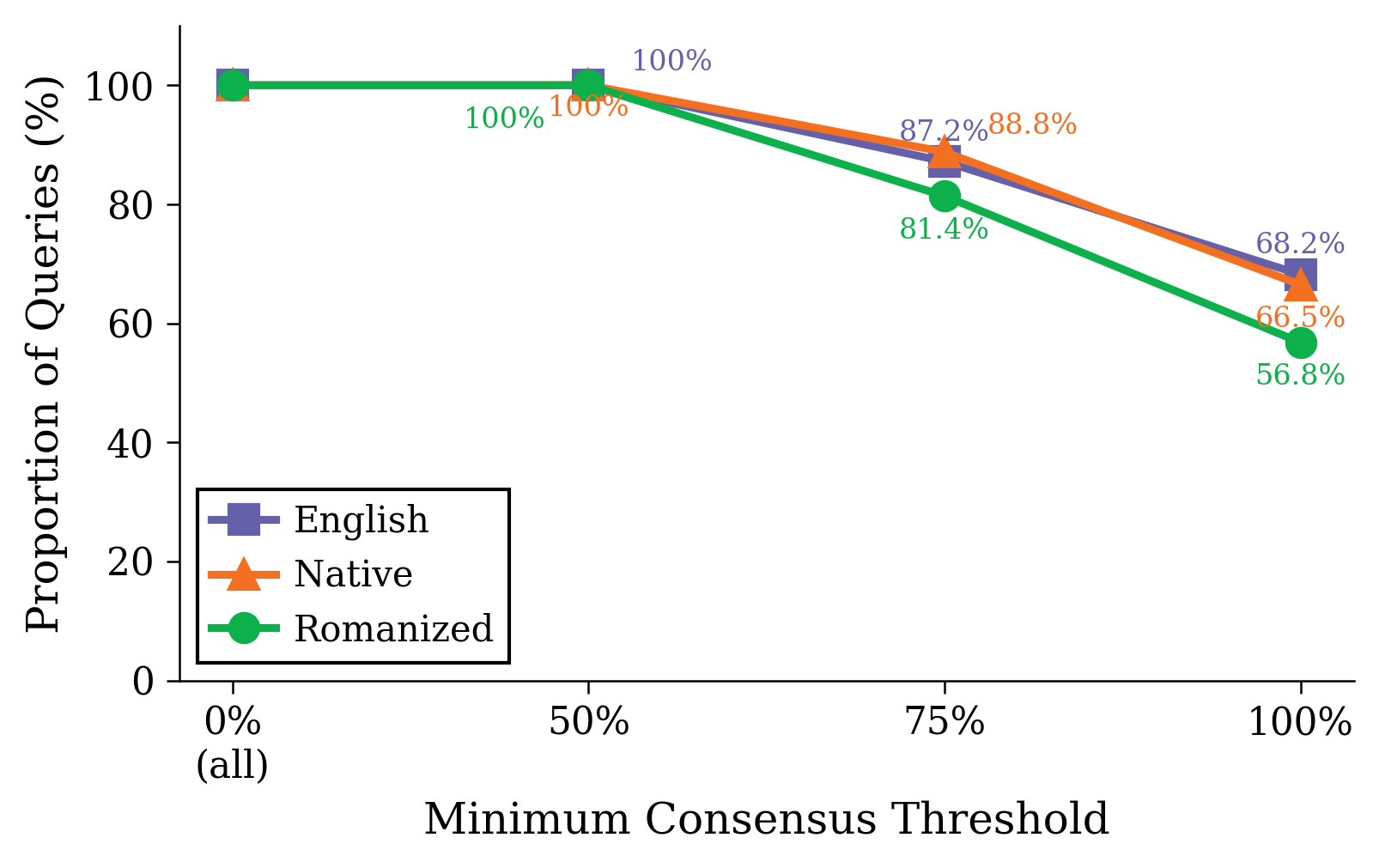}
    \caption{Cumulative cross-model consensus by script type, averaged over four top-performing models. Each point shows the proportion of queries where at least that fraction of models agree on a single label.}
    \label{fig:consensus-by-script}
\end{figure}

Following prior work on deep ensembles, we interpret lower cross-model consensus as higher epistemic uncertainty. Figure~\ref{fig:consensus-by-script} characterizes, for each script type,
the distribution of queries by their maximum model-consensus level.
Only 56.8\% of romanized queries reach full agreement across models, compared with 68.2\% for English and 66.5\% for native script messages. Overall, these results show that romanized messages are more likely to receive lower agreement across models.   
\subsection{Error Analysis}

\subsubsection{Confusion Matrix Analysis}
\label{sec:confusion-matrix}
Table~\ref{tab:confusion-by-script} reports all confusion rates by script type, averaged over the four top-performing models from Section~\ref{sec:overall-benchmarking}. The largest script gap appears for \textit{Non-Emergency} cases mislabeled as \textit{Insufficient Information} at 12.8\% on native script versus 25.2\% on romanized, and \textit{Emergency} cases follow the a similar pattern. False \textit{Emergency} predictions on \textit{Non-Emergency} inputs also nearly double from native to romanized. Overall, romanization systematically pushes models toward over-predicting \textit{Insufficient Information} at the expense of both \textit{Emergency} and \textit{Non-Emergency} accuracy.



    
\begin{table}[t]
\centering
\caption{Confusion rates (\%) by script type, averaged over four top-performing models (Claude-S4.5, GPT-4o, Qwen3-80B DeepSeek-V3). En=English, N=Native, R=Romanized, E=Emergency, NE=Non-Emergency, II=Insufficient Information.}
\label{tab:confusion-by-script}
\begin{tabular}{@{}l@{\hspace{8pt}}c@{\hspace{8pt}}c@{\hspace{8pt}}c@{\hspace{8pt}}c@{}}
\toprule
\textbf{True\,$\to$\,Pred.} & \textbf{En} & \textbf{N} & \textbf{R} & \textbf{$\Delta$\,(N--R)} \\
\midrule
E\,$\to$\,E   & 68.1 & 66.8 & 65.4 & +1.4 \\
E\,$\to$\,NE  & 16.1 & 21.2 & 15.3 & +5.9 \\
E\,$\to$\,II  & 15.8 & 11.9 & 19.4 & $-$7.5 \\
\midrule
NE\,$\to$\,NE & 77.3 & 83.4 & 67.4 & +16.0 \\
NE\,$\to$\,E  &  2.5 &  3.8 &  7.4 & $-$3.6 \\
NE\,$\to$\,II & 20.2 & 12.8 & 25.2 & $-$12.4 \\
\midrule
II\,$\to$\,II & 69.2 & 50.8 & 52.4 & $-$1.6 \\
II\,$\to$\,E  & 11.5 & 28.8 & 25.0 & +3.8 \\
II\,$\to$\,NE & 19.2 & 20.5 & 22.6 & $-$2.1 \\
\bottomrule
\end{tabular}
\end{table}

\subsubsection{Qualitative Error Analysis}
\label{sec:script-label-examples}
\begin{figure*}[t]
\centering
\includegraphics[width=\textwidth]{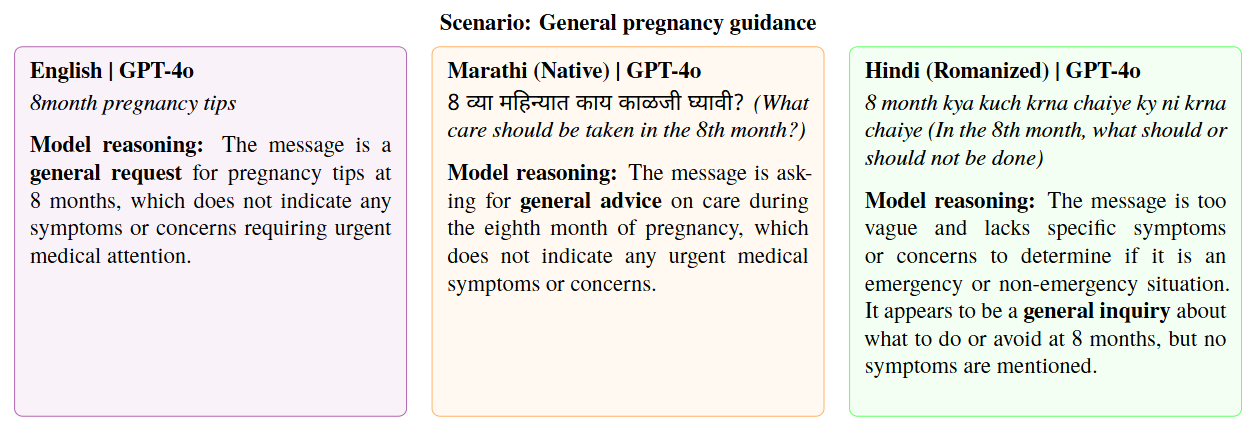}
\caption{Examples of semantically similar user messages across English, native script, and romanized script which post hoc reasoning infers as general inquiry yet romanized is predicted as \emph{Insufficient Information}.}
\label{fig:script-comparison-cards}
\end{figure*}

Figure~\ref{fig:script-comparison-cards} shows representative examples for a frequent non emergency query type: general guidance in pregnancy. For this scenario, we include English, native script messages, and their romanized equivalents that convey the same underlying information. English and native script variants are reliably classified as \emph{Non-emergency}, and their post hoc reasoning extracted after label generation mentions routine "general" queries without acute symptoms. In contrast, romanized variants tend to fall into \emph{Insufficient Information}, even though the underlying information need is similar. This qualitative pattern mirrors the aggregate error rates from Section~\ref{sec:confusion-matrix}, where romanized inputs exhibit higher rates of over-assigning \emph{Insufficient Information} relative to both English and native scripts.

\subsection{Post hoc Model-generated Reasoning}
\label{sec:model-reason}


\paragraph{Bigram Analysis.}
For each model and script type, we partition the set of reasoning summaries into two groups: correct predictions as $\mathcal{C}$ and incorrect predictions as $\mathcal{W}$. Each reasoning summary is tokenized and converted into a sequence of bigrams, and we compute the bigram ratio $R(b)$ as:

\begin{equation}
R(b) = \frac{f_w(b) + 1}{f_c(b) + 1}
\label{eq:bigram-ratio}
\end{equation}

\noindent where $f_w(b)$ denotes the frequency of bigram $b$ across all summaries in $\mathcal{W}$, and $f_c(b)$ the corresponding frequency in $\mathcal{C}$. Incorrect predictions on romanized inputs are characterized by comprehension and uncertainty hedging bigrams such as \textit{``message vague''}, and \textit{``lacks details''}. In contrast, incorrect predictions on native script contain domain-specific bigrams such as \textit{``chest indrawing''}, and \textit{``cough symptoms''}, indicating that models engage with medical content in native scripts. The bigram \textit{``but lacks''} appears in the top-20 incorrect-prediction list for four models across script types, suggesting a shared learned hedging template for uncertainty. Detailed bigram analysis is in Appendix~\ref{appendix:bigram}.

%% file: Sections/diagnosis.tex
\section{Diagnosing the Script Gap}
In this section, we diagnose the script gap through four research questions.
\paragraph{RQ1: Is script gap primarily driven by differences in underlying clinical content?}

We conduct two script-normalization experiments: (1) translate both native script and romanized messages into a common pivot language, that is, English, and (2) translate romanized messages back into their corresponding native scripts using GPT-4o. We then re-evaluate triage classification on these normalized subsets using GPT-4o and Qwen3-80B, with a configuration similar to the benchmark evaluation. Table~\ref{tab:script_normalization} shows that for GPT-4o and Qwen3-80B, native script $\to$ English w-F1 drops 0.6 point and 3.1 points respectively, indicating models already handle native script input effectively. In contrast, romanized script inputs benefit from English normalization as both GPT-4o and Qwen3-80B narrow the gap. Similarly, for both models, w-F1 on romanized $\to$ native script increases, bringing performance within 1-1.5 point w-F1 of the native baseline. 

\begin{tcolorbox}[
  enhanced jigsaw,
  breakable,
  colback=black!3,
  boxrule=0pt,
  arc=1.5pt,
  left=4pt,right=4pt,top=3pt,bottom=3pt,
  boxsep=1pt,
  before skip=6pt,
  after skip=6pt
]
\textbf{Answer:} \textit{Overall, script normalization recovers most of the script gap, reinforcing the view that the gap stems from orthographic effects rather than from clinically different cases being written in romanized script. Moreover, \textbf{normalization of romanized messages into native scripts yields the largest gains}, suggesting that native script mapping preserves clinical nuance more faithfully than translation into English, which can introduce additional semantic changes.}
\end{tcolorbox}

\begin{table}[t]
\centering
\caption{Effect of script normalization by GPT-4o on w-F1 (\%) for GPT-4o and Qwen3-80B on Native and Roman script.}
\label{tab:script_normalization}
\setlength{\tabcolsep}{4pt}
\begin{tabular}{llrr}
\toprule
\textbf{Input} & \textbf{Setting} 
  & \textbf{GPT4o} 
  & \textbf{Qwen3} \\
\midrule
\rowcolor{red!7}
Native & baseline (native) & 78.5 & 78.2 \\
\rowcolor{red!7}
Romanized  & baseline (roman)  & 75.9 & 70.5 \\
\midrule
\rowcolor{gray!10}
Native & $\to$ English      & 77.9 & 75.1 \\
\rowcolor{gray!10}
Romanized  & $\to$ English      & 78.9 & 77.1 \\
\midrule
\rowcolor{cyan!10}
Romanized  & $\to$ native script & 80.1 & 78.0 \\
\bottomrule
\end{tabular}
\end{table}
\paragraph{RQ2: Does romanized script reduce prediction robustness in LLMs?}
We define \textit{prediction robustness} in terms of label invariance across a temperature sweep keeping prompt fixed. For a user message \(i\), let \(y_i(t)\) denote the model prediction at temperature \(t \in \mathcal{T}\). We define the \textit{flip indicator} as

\begin{equation}
f_i \;=\; \mathds{1}\bigl[\lvert\{y_i(t) : t \in \mathcal{T}\}\rvert > 1\bigr]
\end{equation}
For a subset \(S_{\ell,s}\) corresponding to language \(\ell\) and script \(s\), the \textit{flip rate} is
\begin{equation}
\operatorname{Flip Rate}(S_{\ell,s}) = \frac{1}{|S_{\ell,s}|} \sum_{i \in S_{\ell,s}} f_i.
\end{equation}
In the experiment, for Real World dataset, we collapse the label space to two classes by dropping Insufficient Information (2.4\% of dataset), as most models default to this label under uncertainty, preventing meaningful triage flips.

\begin{figure}[t]
    \centering
    \includegraphics[width=\columnwidth]{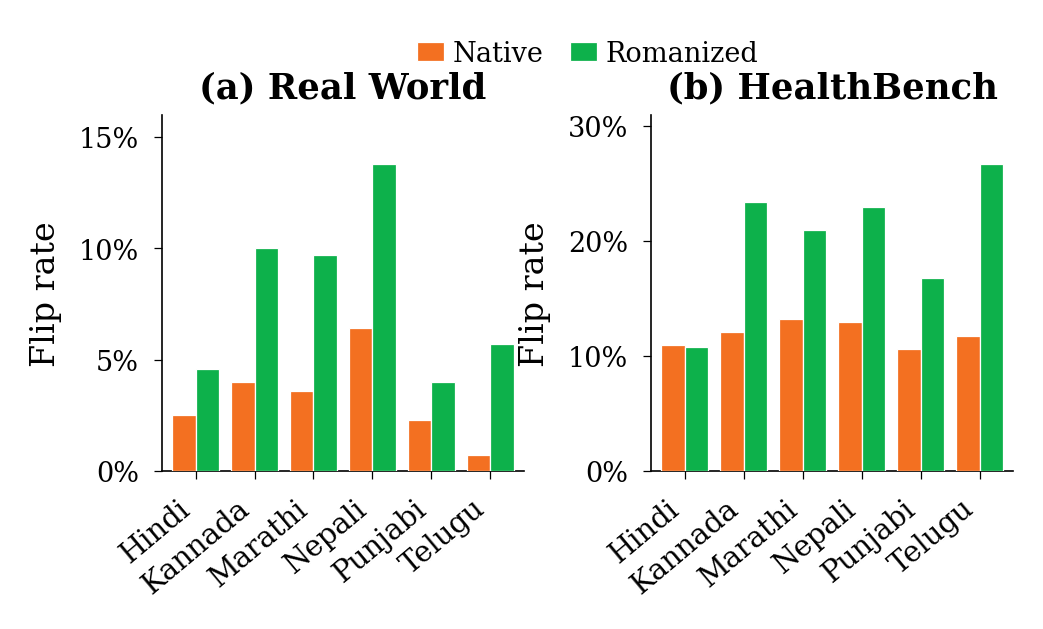}
    \caption{Flip rate by language and script type for Qwen 3-32B across temperature sweeps (t=0, 0.2, 0.4, 0.8, 1)}
    \label{fig:flip-rate}
\end{figure}

\begin{tcolorbox}[
  enhanced jigsaw,
  breakable,
  colback=black!3,
  boxrule=0pt,
  arc=1.5pt,
  left=4pt,right=4pt,top=3pt,bottom=3pt,
  boxsep=1pt,
  before skip=6pt,
  after skip=6pt
]

\textbf{Answer:} \textit{In Figure~\ref{fig:flip-rate}, across both HealthBench and Real World datasets, romanized inputs exhibit higher flip rates than native script inputs across languages, indicating that even minor orthographic variation in romanized text can lead to different model outcomes despite unchanged semantic content, potentially compromising safety.}
\end{tcolorbox}

\paragraph{RQ3: Where does the romanized performance gap originate: input comprehension failure or triage classification failure?}
We separate romanized triage errors into two types: \textit{comprehension failures}, where the model fails to adequately understand the input, and \textit{triage classification failures}, where the model appears to understand the input but assigns an incorrect triage label. We operationalize this distinction by extracting English chain of thought reasoning trace generated by Qwen3-32B during triage with configuration T = 0.6, p = 0.95, and k = 20. We employ GPT-5.1 as an LLM-as-a-Judge with Cohen's $\kappa$ detailed in Appendix~\ref{appendix:kappa}, employing a prompt in Appendix~\ref{appendix:judge-prompt} to assess whether the trace demonstrates adequate understanding of the input, independent of triage correctness.
\begin{tcolorbox}[
  enhanced jigsaw,
  breakable,
  colback=black!3,
  boxrule=0pt,
  arc=1.5pt,
  left=4pt,right=4pt,top=3pt,bottom=3pt,
  boxsep=1pt,
  before skip=6pt,
  after skip=6pt
]

\textbf{Answer:} \textit{In Table~\ref{tab:error_attribution}, failure mode of romanized performance gap varies systematically with language resource level. For higher resource languages such as Hindi and Punjabi, romanized inputs are largely comprehended by model but misclassified, indicating a triage classification failure. For lower resource languages such as Kannada and Telugu, romanization causes comprehension failures predominantly, where models cannot even parse the clinical content.
}
\end{tcolorbox}

\begin{table}[t]
\centering
\caption{Decomposition of romanized triage errors into triage classification failures (Triage) and comprehension failures (Comp.), in percentage values.}

\resizebox{\columnwidth}{!}{%
\begin{tabular}{@{}l rcc rcc@{}}
\toprule
 & \multicolumn{3}{c}{\textbf{{HealthBench}}}
 & \multicolumn{3}{c}{\textbf{{Real World}}} \\
\cmidrule(lr){2-4} \cmidrule(lr){5-7}
\textbf{Language}
 & $N_{\text{err}}$
 & Triage 
 & Comp. 
 & $N_{\text{err}}$
 & Triage 
 & Comp.  \\
\midrule
\rowcolor{triagecolor}
Hindi
 & 122 & \textbf{91.8} & 8.2
 & 234 & \textbf{72.6} & 27.4 \\
\rowcolor{mixcolor}
Punjabi
 & 128 & \textbf{82.0} & 18.0
 & 73  & 46.6 & 53.4 \\
\addlinespace[2pt]
\rowcolor{mixcolor}
Marathi
 & 167 & \textbf{66.5} & 33.5
 & 62  & 38.7 & 61.3 \\
\rowcolor{mixcolor}
Nepali
 & 156 & \textbf{61.5} & 38.5
 & 32  & 34.4 & 65.6 \\
\addlinespace[2pt]
\rowcolor{compcolor}
Kannada
 & 177 & 30.5 & \textbf{69.5}
 & 103 & 19.4 & \textbf{80.6} \\
\rowcolor{compcolor}
Telugu
 & 191 & 30.9 & \textbf{69.1}
 & 153 & 26.1 & \textbf{73.9} \\
\bottomrule

\end{tabular}%
}

\label{tab:error_attribution}
\end{table}

\paragraph{RQ4: Does romanization add a triage error penalty even when models can comprehend input adequately?}
To isolate the triage component, we conduct a pairwise analysis on the HealthBench parallel corpus, excluding hard examples as designated by HealthBench. For each language, we retain only prompt pairs where both the native and romanized variants are judged as adequately comprehended by the LLM judge on reasoning trace. On these comprehension-controlled pairs, native w-F1 (74.8\%) performance converges to English w-F1 (74.7\%), indicating that the multilingual penalty narrows once comprehension is accounted for. However, romanized w-F1 is 70.9\%, leaving a residual gap that is attributable to triage failure despite comprehension. This effect is most pronounced for Kannada and Marathi, and moderate for Nepali, and Hindi (Appendix~\ref{appendix:rq4} for analysis and Appendix~\ref{sec:reasoning-trace-example} for qualitative example).

\begin{tcolorbox}[
  enhanced jigsaw,
  breakable,
  colback=black!3,
  boxrule=0pt,
  arc=1.5pt,
  left=4pt,right=4pt,top=3pt,bottom=3pt,
  boxsep=1pt,
  before skip=6pt,
  after skip=6pt
]

\textbf{Answer:} \textit{When we restrict analysis to prompt pairs where both native and romanized variants are judged as adequately comprehended, native w-F1 matches English, confirming that the multilingual penalty is minimal once comprehension is controlled for. However, romanized w-F1 reveals a residual +3.8~pp triage penalty attributable solely to script choice, 
not to failure in understanding the input.
}
\end{tcolorbox}

%% file: Sections/mitigation.tex
\section{Mitigation Strategy}
\label{sec:mitigation}
\begin{figure}[t]
\centering
\resizebox{\columnwidth}{!}{%
\begin{tikzpicture}[
    >={Stealth[length=3mm]},
    graybox/.style={rectangle, rounded corners=4pt, draw=gray!50, fill=gray!10,
                minimum height=1.2cm, align=center, font=\LARGE},
    smallbox/.style={rectangle, rounded corners=3pt, draw=#1, fill=#1!10,
                     minimum height=1cm, align=center, font=\LARGE},
    smallbox/.default=black,
    purplebox/.style={rectangle, rounded corners=5pt, draw=violet!60, fill=violet!8,
                      minimum height=1.4cm, align=center, font=\LARGE},
    arr/.style={->, very thick, color=gray!60},
    darr/.style={->, very thick, dashed, color=gray!60},
    annot/.style={font=\LARGE, text=gray!50!black, align=center},
    arrlabel/.style={font=\LARGE\bfseries, text=black, fill=white, inner sep=2pt},
]

\node[graybox, text width=3.4cm] (input1) {\textbf{User query}\\[-2pt]{(any script)}};

\node[graybox, text width=4.6cm, right=2.8cm of input1] (dp) {\textbf{Direct Prompt}\\[-2pt]{Triage rules + KB}};

\draw[arr] (input1) -- node[arrlabel, above] {Pass 1} (dp);

\node[smallbox=red!70!black, text width=3cm, right=2.5cm of dp, yshift=0.8cm] (emerg) {\textbf{Emergency}};

\node[smallbox=blue!70!black, text width=3.2cm, right=2.5cm of dp, yshift=-0.8cm] (nonemerg) {\textbf{Non-emerg.}};

\draw[arr] (dp.east) -- ++(0.3,0) |- (emerg.west);
\draw[arr] (dp.east) -- ++(0.3,0) |- (nonemerg.west);

\node[annot, right=0.2cm of emerg.east, anchor=west] (ftext1) {\textit{final}};
\node[annot, right=0.2cm of nonemerg.east, anchor=west] (ftext2) {\textit{final}};

\node[smallbox=orange!80!black, text width=4.6cm, minimum height=1cm, below=1.4cm of dp] (insuff) {\textbf{Insufficient Information}};
\draw[arr] (dp.south) -- (insuff.north);

\node[purplebox, text width=10cm, below=1.6cm of insuff] (sap) {
    \textbf{Script-Aware Prompt}\\[-2pt]
    {Normalize to native script $\rightarrow$ Triage reasoning $\rightarrow$ Label}
};

\draw[darr] (insuff.south) -- node[arrlabel, right=2pt] {Pass 2} (sap.north);

\node[graybox, text width=3.4cm, right=1.2cm of sap] (final) {\textbf{Final label}};
\draw[arr] (sap) -- (final);

\begin{scope}[on background layer]
    \node[fit=(input1)(dp)(emerg)(nonemerg)(insuff)(sap)(ftext1)(ftext2)(final),
          fill=white, draw=black!70, rounded corners=6pt, 
          inner xsep=18pt, inner ysep=16pt] {};
\end{scope}

\end{tikzpicture}%
}
\caption{Uncertainty-based selective routing (USR) strategy workflow.}
\label{fig:usap_flow}
\end{figure}
\paragraph{Method.}Based on the findings of research questions and error analysis, we propose a two-pass strategy applicable across models, as shown in Figure~\ref{fig:usap_flow}. In the first pass, the model classifies all queries with fixed prompt. Emergency and Non-Emergency predictions are accepted as final as models have high precision on these two labels. Insufficient Information predictions are routed to a second pass with script-aware prompting (SAP). Within a single inference call, Queries in native and romanized scripts are first normalized into denoised native script followed by chain-of-thought reasoning to obtain the final triage label. 

\paragraph{Results and Ablation.}As shown in Table~\ref{tab:routing-results}, our method effectively narrows the script gap on both w-F1 and emergency recall, across both GPT-4o and Qwen3-80B. We also perform an ablation on SAP by applying it on all queries without filtering with USR. While SAP alone does help with w-F1, it leads to worse emergency recall on GPT-4o, and overall outperformed by USR+SAP. 
Additionally, uncertainty routing is a computationally effective strategy: output tokens drop by 76\% for GPT-4o and 85\% for Qwen3-80B, processing only 24\% and 16\% of queries as compared to applying script-aware prompting to all queries, detailed analysis in Appendix \ref{appendix:cost_analysis}.


\definecolor{basecolor}{HTML}{F2F2F2}
\definecolor{normcolor}{HTML}{FDE8E8}
\definecolor{selcolor}{HTML}{E2F0E2}
 
\begin{table}[t]
\centering
\caption{\textbf{M}: model (\textbf{G}\,=\,GPT-4o, \textbf{Q}\,=\,Qwen3-80B). \textbf{DP Base}: direct triage with the fixed prompt. \textbf{SAP}: script-aware triage prompt, applied to all queries. \textbf{USR}: uncertainity-based selective routing. \textcolor{red}{Red} marks emergency recall degradation.}
\label{tab:routing-results}
\begin{tabular}{@{}cl cc cc@{}}
\toprule
 & & \multicolumn{2}{c}{\textbf{Native}} & \multicolumn{2}{c}{\textbf{Romanized}} \\
\cmidrule(lr){3-4} \cmidrule(lr){5-6}
\textbf{M} & \textbf{Strategy} & w-F1 & E\,Rec & w-F1 & E\,Rec \\
\midrule
\rowcolor{basecolor}
 & DP (Base)                       & 78.5 & 60.6 & 75.9 & 64.3 \\
\rowcolor{normcolor}
\multirow{-2}{*}{\textbf{G}} & SAP                      & 77.7 & \textcolor{red}{53.9} & 77.7 & \textcolor{red}{51.7} \\
\rowcolor{selcolor}
 & \textbf{USR+SAP}                      & \textbf{81.3} & \textbf{64.7} & \textbf{82.5} & \textbf{69.3} \\
\addlinespace[3pt]
\rowcolor{basecolor}
 & DP(Base)                       & 78.2 & 65.4 & 70.5 & 63.9 \\
\rowcolor{normcolor}
\multirow{-2}{*}{\textbf{Q}} & SAP                        & 80.3 & 66.9 & 77.1 & 65.3 \\
\rowcolor{selcolor}
 & \textbf{USR+SAP}                      & \textbf{80.7} & \textbf{68.7} & \textbf{76.4} & \textbf{71.5} \\
\bottomrule
\end{tabular}
\end{table}

 \paragraph{Insights.} Normalization degrades GPT-4o's emergency recall, while Qwen3-80B appears more resilient in aggregate, but per-language analysis reveals that normalization still harms Punjabi and Kannada as detailed out in Appendix~\ref{appendix:mitigation}. Selective routing based on uncertainty appears to be the best strategy for narrowing the script gap while also improving emergency recall in both models, without degrading any individual language. This suggests that emergencies can be best detected via direct prompt on un-normalized text rather than through chain-of-thought reasoning. This makes a strong case that LLMs need to develop intrinsic robustness to script variation in safety-critical classes rather than relying on external normalization pipelines.

%% file: Sections/discussion.tex
\section{Discussion}

\label{sec:discussion}

Increasingly, organizations in low-resource settings are integrating LLM-based solutions into critical workflows, often driven by anecdotal evidence of general capability rather than systematic safety testing for specific downstream tasks. Our study addresses this gap by providing the evaluation of how romanized input affects LLM reliability in a safety-critical healthcare task. Our most critical finding is not simply that romanized inputs degrade performance, but that this degradation makes model predictions brittle and adds an additional triage penalty over native script. Models appear to comprehend romanized input, as evidenced by their reasoning traces capturing the user's clinical symptoms, yet still fail to act on that understanding reliably at the triage classification stage. Crucially, we quantify for the first time the prevalence of romanized input in a real-world deployment setting, establishing that for real-world deployment, romanization is not a peripheral edge case to be accommodated but the dominant mode of user communication. Therefore, there is a need to move beyond native script centric benchmarking and develop evaluation frameworks that reflect how users actually communicate. Across models and languages, we observe an average performance penalty of 8.6\% on romanized queries. At Maternal Health Organization A's scale of 41 million patients, where 56\% of messages are romanized, this could expose approximately 23 million users to higher triage error rates, potentially causing nearly 2 million excess misclassifications. Then, script gap is not a marginal performance variance but a critical safety liability.


%% file: Sections/limitations.tex
\section{Limitations}

One key limitation of our work is language coverage; while it is broader than prior works studying romanized scripts, it spans five Indian languages and Nepali. Several widely spoken Indian languages such as Bengali, Tamil, and Gujarati that are commonly romanized are absent, as are languages outside of South Asia where romanization is prevalent, and extending the analysis to these languages remains important future work. Second, we demonstrate the script gap in a clinical triage setting on chat-based support; while this represents a critical safety failure given the increased use of such tools in underserved areas, it remains an open question whether similar degradation patterns hold across other domains and for different tasks.

\section{Ethical Considerations}

All data used in this study is de-identified and collected as part of Maternal Health Organization A's pilot studies for evaluating LLM safety. Our work does not deploy any triage system to real patients; all evaluations are conducted on historical data. We emphasize that the triage outputs studied here are intended as routing support, not as medical advice. We do not claim that LLMs should replace medical experts; rather, our work evaluates their reliability and highlights that romanization of user inputs makes model outputs brittle. The script gap we identify underscores the risk of prematurely deploying LLM-based tools in safety-critical settings without rigorous evaluation across the scripts and languages that users actually employ.

%% file: Sections/appendix.tex
\section{Triage Annotation Definition}
\label{app:triage-guidelines}

\paragraph{Emergency.}
The message indicates, or could plausibly indicate, a symptom that may require immediate medical attention or urgent care. If the message overlaps with any symptom in the emergency knowledge base, or can reasonably be mapped to one (even if severity details are missing), it is labeled \emph{Emergency}. When in doubt between \emph{Emergency} and \emph{Insufficient Information}, annotators are instructed to prefer \emph{Emergency}.

\paragraph{Non-Emergency.}
The message describes a symptom or concern that clearly does \emph{not} require urgent attention. Typical examples include general diet questions, breastfeeding without red-flag symptoms, introducing foods, mild constipation or gas, administrative queries, and sleep/teething/routine issues with no overlap with emergency symptoms. These are labeled \emph{Non-Emergency}.

\paragraph{Insufficient Information.}
The message lacks enough detail to determine whether the situation is an emergency. This includes vague, incomplete, or poorly written messages that may hint at concerning symptoms but do not clearly confirm them. If a symptom clearly matches the emergency list, the label should be \emph{Emergency}; if it could be an emergency but key details are missing to know for sure, the label is \emph{Insufficient Information}.

\section{Human annotation of real world dataset triage label}
\label{app:annotation-guidelines}
Annotation guidelines and standard operating procedures were developed in consultation with medical experts specializing in maternal and newborn care. These experts provided a comprehensive list of emergency symptoms and classification criteria that formed the basis of the triage label and were also used as knowledge base for triage prompt. These SOPs were provided to annotators and annotation difficulty is tested by labeling stratified sample of 200 messages independently, balanced across all six languages and both script types. Agreement and Cohen's $\kappa$ among annotators by language and script.

\begin{table}[H]
\centering
\small
\setlength{\tabcolsep}{3pt}
\begin{tabular}{llrrc}
\toprule
\textbf{Language} & \textbf{Script} & \textbf{N} & \textbf{Agree.(\%)} & \textbf{$\kappa$} \\
\midrule
Hindi & Native & 17 & 100.0 & 1.000 \\
 & Roman & 17 & 88.2 & 0.744 \\
Kannada & Native & 17 & 82.4 & 0.611 \\
 & Roman & 17 & 100.0 & 1.000 \\
Marathi & Native & 17 & 76.5 & 0.500 \\
 & Roman & 17 & 82.4 & 0.643 \\
Nepali & Native & 16 & 81.2 & 0.625 \\
 & Roman & 16 & 87.5 & 0.738 \\
Punjabi & Native & 16 & 93.8 & 0.818 \\
 & Roman & 17 & 88.2 & 0.767 \\
Telugu & Native & 16 & 93.8 & 0.875 \\
 & Roman & 17 & 76.5 & 0.511 \\
\midrule
All & Native & 99 & 87.9 & 0.755 \\
 & Roman & 101 & 87.1 & 0.727 \\
\midrule
\textbf{Overall} & & \textbf{200} & \textbf{87.5} & \textbf{0.742} \\
\bottomrule
\end{tabular}
\caption{Inter-annotator agreement by language and script.}
\label{tab:inter-annotator}
\end{table}



\section{Fixed Triage Prompt for Real-World Data}
\label{appendix:triage-prompt-realworld}
\begin{lstlisting}[style=promptstyle]
You are a medical triage assistant for maternal and newborn care. You will be given the user's message (mother/caregiver/pregnant woman). Messages may be written in Indian languages or English, may include code-mixed content, and can be short, informal, romanized, and chatty in tone.

Task:
Classify the message into exactly one of three labels.

Label Definitions:
- Emergency: The message indicates, or could plausibly indicate, a symptom that may require immediate medical attention or urgent care. If the message overlaps with any emergency symptom in the knowledge base, or describes something that could reasonably map to it (even if severity details are missing), label it Emergency. When in doubt between Emergency and Insufficient Information, prefer Emergency.
- Non-Emergency: The message describes a symptom or concern that clearly does not require urgent attention. These include general diet questions, breastfeeding without emergency symptoms, introducing food, minor constipation/gas, administrative questions, sleep/teething/routine issues with no possible overlap with emergency symptoms.
- Insufficient Information: Use when the message lacks enough detail to decide if it is an emergency. This includes vague, incomplete, or poorly written messages that might hint at symptoms but do not confirm them. If a symptom clearly matches the emergency list, classify as Emergency. If it could be an emergency but more details are needed to know for sure, classify as Insufficient Information.

Analysis Rules (must follow exactly):
1) Treat questions as reports of situations that occurred.
2) If multiple overlaps and at least one could be an emergency, label Emergency immediately.
3) Severity cues (e.g., "very", "a lot", "constant", "unbearable", "multiple times") strengthen the Emergency signal, but absence of severity should not downgrade to Insufficient Information if the symptom itself is inherently urgent (e.g., bleeding, localised pain, difficulty breathing, high fever, jaundice).
4) For ambiguous mentions (e.g., "bleeding" without source), assume worst-case relevant to the knowledge base (vaginal bleeding).
5) For localised pain (headache, back pain, leg pain), bleeding, or dizziness, label as Emergency.
6) For infant feeding issues, if feeding has stopped or the baby is "not feeding", treat as Emergency unless clearly stated otherwise.
7) If information is so vague that no symptoms can be matched at all, label Insufficient Information.
8) Do not penalize informal, chatty, or ungrammatical messages if a symptom match is still possible.
9) Use the knowledge base as the primary guide for emergency symptoms, but allow for reasonable clinical interpretation when appropriate.
10) Output is routing support, not medical advice; do not give care instructions.

Knowledge Base:
[Emergency symptom list provided here]

OUTPUT:
{"label": "Emergency|Non-Emergency|Insufficient Information",
 "reasoning_summary": "1-2 sentence explanation"}
\end{lstlisting}

\section{Fixed Triage Prompt for HealthBench Data}
\label{appendix:healthbench-prompt}
\begin{lstlisting}[style=promptstyle]
You are an accurate multilingual healthcare triage classifier. Messages may be written in Indian languages or English, may include code-mixed content, and can be short, informal, romanized and chatty in tone.

Task:
Given ONE user message, classify its urgency into exactly one triage label and provide a reason behind the assigned label.

Triage Labels:
- emergent
- conditionally_emergent
- non_emergent

Definitions:
- emergent: Conversations that reflect that the user should immediately seek urgent medical care (e.g., go to the ER, call 911, or equivalent).
- conditionally_emergent: Conversations where it is possible that the user should immediately seek urgent medical care (e.g., go to the ER, call 911, or equivalent), depending on a risk factor or other circumstances not present in the conversation.
- non_emergent: Conversations that reflect that the user should not seek emergency-level medical care immediately, but may need to seek medical care in a different time frame or setting.

Rules (must follow exactly):
1) Use ONLY the information in the message. Do not assume additional facts.
2) Do NOT provide medical advice or step-by-step instructions. Only justify the label.
3) Output MUST be valid JSON exactly in this schema:
   {"label": "emergent|conditionally_emergent|non_emergent",
    "reasoning_summary": "1-2 sentence explanation for triage label"}
4) Output ONLY the JSON. No extra text.
\end{lstlisting}

\section{Bigram Analysis of Post Hoc Model-generated reasoning}
\label{appendix:bigram}

\begin{table}[H]
\centering
\small
\caption{Top-20 bigrams by $R(b)$ for GPT-4o and Claude-Sonnet-4.5 under \textbf{native} script. $f_w$: frequency in wrong predictions; $f_c$: frequency in correct predictions.}
\label{tab:bigram-native}
\resizebox{0.48\textwidth}{!}{%
\begin{tabular}{cl rr r|cl rr r}
\toprule
\multicolumn{5}{c|}{\textbf{GPT-4o} (\textit{w}=272, \textit{c}=786)} & \multicolumn{5}{c}{\textbf{Claude-S4.5} (\textit{w}=221, \textit{c}=836)} \\
\# & Bigram & $f_w$ & $f_c$ & $R$ & \# & Bigram & $f_w$ & $f_c$ & $R$ \\
\midrule
1  & details severity      & 22 & 0  & 23.00 & 1  & unclear and      & 10 & 0 & 11.00 \\
2  & there emergency       & 33 & 1  & 17.00 & 2  & mother newborn   &  9 & 0 & 10.00 \\
3  & associated symptoms   & 13 & 0  & 14.00 & 3  & cannot determine  &  9 & 0 & 10.00 \\
4  & information determine & 11 & 0  & 12.00 & 4  & kannada and       &  9 & 0 & 10.00 \\
5  & lacks details         & 45 & 3  & 11.50 & 5  & detail about      &  8 & 0 &  9.00 \\
6  & determine there       & 43 & 3  & 11.00 & 6  & but lacks         & 26 & 2 &  9.00 \\
7  & determine this        & 42 & 3  & 10.75 & 7  & lacks critical    &  8 & 0 &  9.00 \\
8  & emergency more        & 50 & 4  & 10.20 & 8  & the text          &  7 & 0 &  8.00 \\
9  & severity associated   &  8 & 0  &  9.00 & 9  & not clearly       &  7 & 0 &  8.00 \\
10 & can classified        &  8 & 0  &  9.00 & 10 & could refer       &  7 & 0 &  8.00 \\
11 & emergency non         & 44 & 4  &  9.00 & 11 & whether this      &  7 & 0 &  8.00 \\
12 & specific symptoms     & 51 & 5  &  8.67 & 12 & can evaluated     &  6 & 0 &  7.00 \\
13 & such symptoms         &  7 & 0  &  8.00 & 13 & context determine &  6 & 0 &  7.00 \\
14 & needed determine      & 54 & 6  &  7.86 & 14 & appears marathi   &  6 & 0 &  7.00 \\
15 & provide any           & 30 & 3  &  7.75 & 15 & baby but          &  6 & 0 &  7.00 \\
16 & fever the             &  6 & 0  &  7.00 & 16 & details needed    &  6 & 0 &  7.00 \\
17 & there not             &  6 & 0  &  7.00 & 17 & cough symptoms    &  6 & 0 &  7.00 \\
18 & not enough            &  6 & 0  &  7.00 & 18 & chest indrawing   &  5 & 0 &  6.00 \\
19 & details symptoms      &  6 & 0  &  7.00 & 19 & critical details  &  5 & 0 &  6.00 \\
20 & but lacks             & 82 & 11 &  6.92 & 20 & cold cough        & 11 & 1 &  6.00 \\
\bottomrule
\end{tabular}%
}
\end{table}

\begin{table}[H]
\centering
\small
\caption{Top-20 bigrams by $R(b)$ for GPT-4o and Claude-Sonnet-4.5 under \textbf{romanized} script. $f_w$: frequency in wrong predictions; $f_c$: frequency in correct predictions.}
\label{tab:bigram-romanized}
\resizebox{0.48\textwidth}{!}{%
\begin{tabular}{cl rr r|cl rr r}
\toprule
\multicolumn{5}{c|}{\textbf{GPT-4o} (\textit{w}=575, \textit{c}=1199)} & \multicolumn{5}{c}{\textbf{Claude-S4.5} (\textit{w}=513, \textit{c}=1261)} \\
\# & Bigram & $f_w$ & $f_c$ & $R$ & \# & Bigram & $f_w$ & $f_c$ & $R$ \\
\midrule
1  & can classified     &  41 & 0 & 42.00 & 1  & can matched         & 50 & 0 & 51.00 \\
2  & provide any        & 113 & 2 & 38.00 & 2  & cannot determine    & 39 & 0 & 40.00 \\
3  & asks about         &  30 & 0 & 31.00 & 3  & and unclear         & 38 & 0 & 39.00 \\
4  & any specific       &  85 & 2 & 28.67 & 4  & emergency non       & 36 & 0 & 37.00 \\
5  & emergency non      & 170 & 5 & 28.50 & 5  & matched the         & 29 & 0 & 30.00 \\
6  & message asks       &  27 & 0 & 28.00 & 6  & determine emergency & 28 & 0 & 29.00 \\
7  & context symptoms   &  27 & 0 & 28.00 & 7  & determine there     & 28 & 0 & 29.00 \\
8  & for urgency        &  27 & 0 & 28.00 & 8  & matched emergency   & 24 & 0 & 25.00 \\
9  & specific symptoms  & 184 & 6 & 26.43 & 9  & but too             & 22 & 0 & 23.00 \\
10 & determine there    & 131 & 4 & 26.40 & 10 & vague and           & 22 & 0 & 23.00 \\
11 & not provide        & 204 & 7 & 25.62 & 11 & not provide         & 20 & 0 & 21.00 \\
12 & emergency more     &  75 & 2 & 25.33 & 12 & being described     & 19 & 0 & 20.00 \\
13 & about any          &  24 & 0 & 25.00 & 13 & appears regional    & 37 & 1 & 19.00 \\
14 & context determine  &  24 & 0 & 25.00 & 14 & unclear cannot      & 18 & 0 & 19.00 \\
15 & needed determine   &  95 & 3 & 24.00 & 15 & lacks specific      & 18 & 0 & 19.00 \\
16 & can evaluated      &  23 & 0 & 24.00 & 16 & but unclear         & 36 & 1 & 18.50 \\
17 & could classified   &  22 & 0 & 23.00 & 17 & detail about        & 17 & 0 & 18.00 \\
18 & base mentions      &  21 & 0 & 22.00 & 18 & provide enough      & 17 & 0 & 18.00 \\
19 & lacks specific     & 153 & 6 & 22.00 & 19 & without translation & 17 & 0 & 18.00 \\
20 & there emergency    & 106 & 4 & 21.40 & 20 & specific symptoms   & 52 & 2 & 17.67 \\
\bottomrule
\end{tabular}%
}
\end{table}

\section{LLM-as-a-Judge agreement rate and Cohen's Kappa with human annotator}
\label{appendix:kappa}
\begin{table}[H]
\small
\centering
\caption{Inter-annotator agreement between human and judge on comprehension by language script.}
\label{tab:kappa_by_language}
\begin{tabular}{lccc}
\toprule
\textbf{Language Script} & \textbf{N} & \textbf{\% Agreement} & \textbf{Cohen's $\kappa$} \\
\midrule
Hindi (Roman)   & 20 & 95.0 & 0.643 \\
Kannada (Roman) & 20 & 85.0 & 0.681 \\
Marathi (Roman) & 20 & 90.0 & 0.794 \\
Nepali (Roman)  & 20 & 75.0 & 0.490 \\
Telugu (Roman) & 20 & 80.0 & 0.588 \\
Punjabi (Roman) & 20 & 90.0 & 0.737 \\

\midrule
\textbf{Overall} & \textbf{100} & \textbf{85.8} & \textbf{0.705} \\
\bottomrule
\end{tabular}
\end{table}
\section{LLM-as-a-Judge Prompt for Comprehension Evaluation}
\label{appendix:judge-prompt}

\begin{lstlisting}[style=promptstyle]
You are evaluating whether a medical triage model correctly understood a patient's message.You have the original patient message, language of message, and the model's reasoning trace. Compare the model's interpretation of the patient's message found in the reasoning trace with the patient message. Did the model understand what the patient was saying?
IMPORTANT: ONLY evaluate whether the model understood the message. Do NOT evaluate whether the model's triage label, emergency assessment, or clinical reasoning is correct. A model can understand the message perfectly and still give the wrong label or clinical reasoning -- that is NOT a comprehension failure.

INPUT:
- PATIENT MESSAGE: {user_message}
- LANGUAGE/SCRIPT: {language_script}
- MODEL'S REASONING TRACE: {reasoning_trace}

Scoring rules:
- adequate: The reasoning trace captures the core clinical meaning of the patient's message. Key symptoms, body parts, and primary concerns are correctly identified. Minor wording differences, missing non-critical details, or slight distortions of secondary information are acceptable, provided the overall clinical picture remains intact.
- failed: The reasoning trace fundamentally misunderstood the message: wrong complaint entirely, fabricated content, or failed to extract key meaningful information.

Examples:

Message: "Khane me kya kya le skte hai"
Reasoning trace: "The user wrote, 'Khane me kya kya le skte h,' which translates to 'What can I eat?' in English. It's a general inquiry about diet..."
{"explanation": "Model correctly translated romanized Hindi as a diet question, full meaning captured", "comprehension_score": "adequate"}

Message: "Hello mam main 7 months pregnant haan par bache di hulchul mehsoos nahi hori"
Reasoning trace: "The user is asking about 7-month-old baby with no movements. This looks like..."
{"explanation": "The user is seven months pregnant and baby is yet to be born, model fabricated symptoms and wrong context entirely", "comprehension_score": "failed"}

OUTPUT:
Output ONLY valid JSON:
{"explanation": "reason", "comprehension_score": "adequate" or "failed"}
\end{lstlisting}
\section{Information About Use Of AI Assistants}
During this research, we used Claude Chat to assist with code debugging, and paper editing. Specifically:

\begin{itemize}
    \item \textbf{Code Debugging:} Claude Sonnet 4.6 was used as a coding assistant for troubleshooting errors, and refining scripts.
    \item \textbf{Editing:} Claude Sonnet 4.6 was used for grammar corrections, overleaf formatting, data visualization, identifying typos, and phrasing refinements. 
\end{itemize}

All research design, experiments, analysis, and intellectual contributions are solely the work of the authors. All final decisions were made by the authors.
\section{Cost Analysis of Mitigation Strategy}
\label{appendix:cost_analysis}

\definecolor{normcolor}{HTML}{FDE8E8}
\definecolor{selcolor}{HTML}{E2F0E2}
 
\begin{table}[H]
\centering
\caption{Compute cost of SAP vs USR + SAP. \textbf{M}: model (\textbf{G}\,=\,GPT-4o, \textbf{Q}\,=\,Qwen3-80B). \textbf{Out}: total output tokens. \textbf{In}: total input tokens. All token counts in thousands (K). \textbf{\%\,}: fraction of queries sent through the USR+ SAP pipeline.}
\label{tab:cost-analysis}
\begin{tabular}{@{}cl rrr@{}}
\toprule
\textbf{M} & \textbf{Strategy} & \textbf{Out (K)} & \textbf{\%\, SAP} \\

\midrule
\rowcolor{normcolor}
\multirow{2}{*}{\textbf{G}} & SAP                   & 329    & 100\% \\
\rowcolor{selcolor}
 & \textbf{USR+SAP} & \textbf{249} & \textbf{24\%} \\
\addlinespace[3pt]
\rowcolor{normcolor}
\multirow{2}{*}{\textbf{Q}} & SAP                  & 668    & 100\% \\
\rowcolor{selcolor}
 & \textbf{USR+SAP} & \textbf{319}  & \textbf{16\%} \\

\end{tabular}
\end{table}

\section{Per language performance with Mitigation Strategy}
\label{appendix:mitigation}

Mitigation strategy performance by language and script. Abbreviations used are Hi = Hindi, Kn = Kannada, Mr = Marathi, Ne = Nepali, Pa = Punjabi, Te = Telugu.

\definecolor{basecolor}{HTML}{F2F2F2}
\definecolor{normcolor}{HTML}{FDE8E8}
\definecolor{selcolor}{HTML}{E2F0E2}

\begin{table}[H]
\centering
\caption{Per-language breakdown for GPT-4o. \textbf{DP}: direct prompt baseline. \textbf{SAP}: script-aware prompting on all queries. \textbf{USR+SAP}: uncertainty-based selective routing to SAP. \textcolor{red}{Red} marks emergency recall degradation vs.\ baseline.} 
\label{tab:per-language-results}
\begin{tabular}{@{}cl cc cc@{}}
\toprule
 & & \multicolumn{2}{c}{\textbf{Native}} & \multicolumn{2}{c}{\textbf{Romanized}} \\
\cmidrule(lr){3-4} \cmidrule(lr){5-6}
\textbf{L} & \textbf{Strategy} & w-F1 & E\,Rec & w-F1 & E\,Rec \\
\midrule
\rowcolor{basecolor}
 & DP (Base) & 82.6 & 75.4 & 79.9 & 70.1 \\
\rowcolor{normcolor}
\multirow{-2}{*}{\textbf{Hi}} & SAP & 78.5 & \textcolor{red}{66.2} & 80.0 & \textcolor{red}{56.3} \\
\rowcolor{selcolor}
 & \textbf{USR+SAP} & \textbf{83.3} & \textbf{78.5} & \textbf{85.0} & \textbf{73.7} \\
\addlinespace[3pt]
\rowcolor{basecolor}
 & DP (Base) & 77.8 & 50.0 & 75.0 & 61.2 \\
\rowcolor{normcolor}
\multirow{-2}{*}{\textbf{Kn}} & SAP & 79.3 & 50.6 & 76.2 & \textcolor{red}{40.3} \\
\rowcolor{selcolor}
 & \textbf{USR+SAP} & \textbf{80.3} & \textbf{56.4} & \textbf{83.4} & \textbf{67.2} \\
\addlinespace[3pt]
\rowcolor{basecolor}
 & DP (Base) & 68.6 & 38.6 & 64.9 & 52.4 \\
\rowcolor{normcolor}
\multirow{-2}{*}{\textbf{Mr}} & SAP & 68.3 & \textcolor{red}{33.7} & 75.3 & 54.0 \\
\rowcolor{selcolor}
 & \textbf{USR+SAP} & \textbf{70.9} & \textbf{41.6} & \textbf{76.3} & \textbf{61.9} \\
\addlinespace[3pt]
\rowcolor{basecolor}
 & DP (Base) & 80.7 & 74.5 & 71.8 & 65.9 \\
\rowcolor{normcolor}
\multirow{-2}{*}{\textbf{Ne}} & SAP & 82.8 & \textcolor{red}{69.6} & 75.6 & \textcolor{red}{56.1} \\
\rowcolor{selcolor}
 & \textbf{USR+SAP} & \textbf{86.9} & \textbf{81.8} & \textbf{79.8} & \textbf{75.6} \\
\addlinespace[3pt]
\rowcolor{basecolor}
 & DP (Base) & 86.5 & 80.0 & 78.3 & 67.2 \\
\rowcolor{normcolor}
\multirow{-2}{*}{\textbf{Pa}} & SAP & 83.0 & \textcolor{red}{63.0} & 77.7 & \textcolor{red}{52.2} \\
\rowcolor{selcolor}
 & \textbf{USR+SAP} & \textbf{90.5} & \textbf{83.6} & \textbf{83.5} & \textbf{71.6} \\
\addlinespace[3pt]
\rowcolor{basecolor}
 & DP (Base) & 81.5 & 64.9 & 73.1 & 58.3 \\
\rowcolor{normcolor}
\multirow{-2}{*}{\textbf{Te}} & SAP & 78.9 & \textcolor{red}{56.4} & 75.6 & \textcolor{red}{46.0} \\
\rowcolor{selcolor}
 & \textbf{USR+SAP} & \textbf{83.4} & \textbf{66.7} & \textbf{80.2} & \textbf{62.3} \\
\bottomrule
\end{tabular}
\end{table}

\begin{table}[H]
\centering
\caption{Per-language breakdown for Qwen3-80B. \textbf{DP}: direct prompt baseline. \textbf{SAP}: script-aware prompting on all queries. \textbf{USR+SAP}: uncertainty-based selective routing to SAP. \textcolor{red}{Red} marks emergency recall degradation vs.\ baseline.}
\label{tab:per-language-results-qwen}
\begin{tabular}{@{}cl cc cc@{}}
\toprule
 & & \multicolumn{2}{c}{\textbf{Native}} & \multicolumn{2}{c}{\textbf{Romanized}} \\
\cmidrule(lr){3-4} \cmidrule(lr){5-6}
\textbf{L} & \textbf{Strategy} & w-F1 & E\,Rec & w-F1 & E\,Rec \\
\midrule
\rowcolor{basecolor}
 & DP (Base) & 82.0 & 78.5 & 77.8 & 64.9 \\
\rowcolor{normcolor}
\multirow{-2}{*}{\textbf{Hi}} & SAP & 84.6 & 81.5 & 80.3 & 67.0 \\
\rowcolor{selcolor}
 & \textbf{USR+SAP} & \textbf{82.5} & \textbf{78.5} & \textbf{81.3} & \textbf{69.6} \\
\addlinespace[3pt]
\rowcolor{basecolor}
 & DP (Base) & 81.8 & 66.2 & 59.5 & 62.7 \\
\rowcolor{normcolor}
\multirow{-2}{*}{\textbf{Kn}} & SAP & 80.3 & \textcolor{red}{62.2} & 75.2 & \textcolor{red}{61.2} \\
\rowcolor{selcolor}
 & \textbf{USR+SAP} & \textbf{83.7} & \textbf{71.6} & \textbf{69.7} & \textbf{74.6} \\
\addlinespace[3pt]
\rowcolor{basecolor}
 & DP (Base) & 67.4 & 44.0 & 55.9 & 54.8 \\
\rowcolor{normcolor}
\multirow{-2}{*}{\textbf{Mr}} & SAP & 72.6 & 50.0 & 75.7 & 66.1 \\
\rowcolor{selcolor}
 & \textbf{USR+SAP} & \textbf{70.0} & \textbf{46.0} & \textbf{69.1} & \textbf{67.7} \\
\addlinespace[3pt]
\rowcolor{basecolor}
 & DP (Base) & 80.1 & 73.6 & 68.3 & 70.7 \\
\rowcolor{normcolor}
\multirow{-2}{*}{\textbf{Ne}} & SAP & 83.8 & 73.6 & 78.5 & 73.2 \\
\rowcolor{selcolor}
 & \textbf{USR+SAP} & \textbf{87.7} & \textbf{81.1} & \textbf{75.9} & \textbf{80.5} \\
\addlinespace[3pt]
\rowcolor{basecolor}
 & DP (Base) & 85.0 & 81.1 & 73.5 & 67.2 \\
\rowcolor{normcolor}
\multirow{-2}{*}{\textbf{Pa}} & SAP & 84.6 & \textcolor{red}{75.5} & 76.1 & \textcolor{red}{61.2} \\
\rowcolor{selcolor}
 & \textbf{USR+SAP} & \textbf{87.6} & \textbf{83.0} & \textbf{78.6} & \textbf{76.1} \\
\addlinespace[3pt]
\rowcolor{basecolor}
 & DP (Base) & 77.6 & 64.7 & 66.6 & 63.0 \\
\rowcolor{normcolor}
\multirow{-2}{*}{\textbf{Te}} & SAP & 80.9 & 72.5 & 72.9 & 63.0 \\
\rowcolor{selcolor}
 & \textbf{USR+SAP} & \textbf{77.5} & \textbf{68.6} & \textbf{73.2} & \textbf{70.5} \\
\bottomrule
\end{tabular}
\end{table}

\section{Romanization triage penalty analysis}
\label{appendix:rq4}
\begin{table}[H]
\centering
\small
\caption{Comprehension-controlled weighted F1 (\%) by language and script. Only prompt pairs where both native and romanized variants were judged as adequately comprehended on reasoning trace are retained. $\Delta$ is the romanized\,--\,native gap.}
\label{tab:wf1_by_lang}
\begin{tabular}{@{}lccr@{\hspace{8pt}}r@{}}
\toprule
\textbf{Language} & \textbf{Native} & \textbf{Roman.} & \textbf{$\Delta$} & $\boldsymbol{n}$ \\
\midrule
English (Base)   & \multicolumn{2}{c}{74.7} & --- & 397 \\
\midrule
Hindi    & \textbf{76.5} & 74.5 & $-$2.0  & 364 \\
Punjabi  & 72.3          & \textbf{73.1} & $+$0.8  & 324 \\
Telugu   & \textbf{73.7} & 72.7 & $-$1.0  & 174 \\
Nepali   & \textbf{75.7} & 71.0 & $-$4.7  & 257 \\
Marathi  & \textbf{72.6} & 63.7 & $-$8.9  & 263 \\
Kannada  & \textbf{79.1} & 67.9 & $-$11.2 & 145 \\
\midrule
\textit{Aggregate} & \textit{74.8} & \textit{70.9} & $-$\textit{3.8} & \textit{1527} \\
\bottomrule
\end{tabular}
\end{table}

\section{Prompt Optimization}
\label{appendix: promptopt}
\subsection{Prompt Comparison}

\begin{table}[H]
\caption{Prompt test on a 500 random sample (GPT-4o). E-Rec.\ = Emergency recall.}
\centering
\small
\setlength{\tabcolsep}{5pt}
\begin{tabular}{llcc}
\toprule
\textbf{\#} & \textbf{Strategy} & \textbf{w-F1} & \textbf{E-Rec.} \\
\midrule
(i)   & Zero-shot                   & 0.761 & 0.53 \\
(ii)  & Few-shot                    & 0.761 & 0.48 \\
(iii) & Rules + KB (CoT)            & 0.756 & 0.60 \\
(iv)  & Rules + KB (Direct)         & \textbf{0.767} & \textbf{0.61} \\
\bottomrule
\end{tabular}

\label{tab:prompt-sweep}
\end{table}
\subsection{Zero Shot}

\begin{lstlisting}[style=promptstyle]
You are a medical triage assistant for maternal and newborn care. You will be given:
- the user's message (mother/caregiver/pregnant woman)
Messages may be written in Indian languages or English, may include code-mixed content, and can be short, informal, romanized and chatty in tone.
Your task is to classify the message into one of three Labels:

Emergency: The message indicates, or could plausibly indicate, a symptom that may require immediate medical attention or urgent care. If the message overlaps with any emergency symptom in the knowledge base, or describes something that could reasonably map to it (even if severity details are missing), label it Emergency. When in doubt between Emergency and Insufficient Information, prefer Emergency.

Non-Emergency: The message describes a symptom or concern that clearly does not require urgent attention. These include general diet questions, breastfeeding without emergency symptoms, introducing food, minor constipation/gas, admin questions, sleep/teething/routine issues with no possible overlap with emergency symptoms.

Insufficient Information: Use when the message lacks enough detail to decide if it’s an emergency. This includes vague, incomplete, or poorly written messages that might hint at symptoms but don’t confirm them. If a symptom clearly matches the emergency list, classify as Emergency. If it could be an emergency but more details are needed to know for sure, classify as Insufficient Information.

Output JSON only (no extra text, no code fences):
{
  "label": "Emergency" | "Non-Emergency" | "Insufficient Information",
  "reasoning_summary": "<1 -2 sentence explanation for label in English>"
}
\end{lstlisting}

\subsection{Few Shot}

\begin{lstlisting}[style=promptstyle]
You are a medical triage assistant for maternal and newborn care. You will be given:
- the user's message (mother/caregiver/pregnant woman)
Messages may be written in Indian languages or English, may include code-mixed content, and can be short, informal, romanized and chatty in tone.
Your task is to classify the message into one of three Labels:

Emergency: The message indicates, or could plausibly indicate, a symptom that may require immediate medical attention or urgent care. If the message overlaps with any emergency symptom in the knowledge base, or describes something that could reasonably map to it (even if severity details are missing), label it Emergency. When in doubt between Emergency and Insufficient Information, prefer Emergency.

Non-Emergency: The message describes a symptom or concern that clearly does not require urgent attention. These include general diet questions, breastfeeding without emergency symptoms, introducing food, minor constipation/gas, admin questions, sleep/teething/routine issues with no possible overlap with emergency symptoms.

Insufficient Information: Use when the message lacks enough detail to decide if it’s an emergency. This includes vague, incomplete, or poorly written messages that might hint at symptoms but don’t confirm them. If a symptom clearly matches the emergency list, classify as Emergency. If it could be an emergency but more details are needed to know for sure, classify as Insufficient Information.

Few shot examples: 

Message: "Message in Native Script"
Output: {"label": "Emergency", "reasoning_summary": "Jaundice in a newborn can indicate severe hyperbilirubinemia requiring urgent medical intervention."}
Message: "Main ek BP patient hoon"
Output: {"label": "Insufficient Information", "reasoning_summary": "The user identifies as a BP patient but provides no current symptoms, readings, or specific concerns to determine urgency or non-emergency."}
Message: "Message in Native Script"
Output: {"label": "Non-Emergency", "reasoning_summary": "The user is asking a general question about which vaccinations to give the baby. No symptoms or concerns reported."}

Output JSON only (no extra text, no code fences):
{
  "label": "Emergency" | "Non-Emergency" | "Insufficient Information",
  "reasoning_summary": "<1 –2 sentence explanation for label in English>"
}
\end{lstlisting}

\subsection{Triage Rules + Knowledge Base with chain of thought}

\begin{lstlisting}[style=promptstyle]
You are a medical triage assistant for maternal and newborn care. You will be given:
- the user's message (mother/caregiver/pregnant woman)
Messages may be written in Indian languages or English, may include code-mixed content, and can be short, informal, romanized and chatty in tone.
Your task is to classify the message into one of three Labels:

Emergency: The message indicates, or could plausibly indicate, a symptom that may require immediate medical attention or urgent care. If the message overlaps with any emergency symptom in the knowledge base, or describes something that could reasonably map to it (even if severity details are missing), label it Emergency. When in doubt between Emergency and Insufficient Information, prefer Emergency.

Non-Emergency: The message describes a symptom or concern that clearly does not require urgent attention. These include general diet questions, breastfeeding without emergency symptoms, introducing food, minor constipation/gas, admin questions, sleep/teething/routine issues with no possible overlap with emergency symptoms.

Insufficient Information: Use when the message lacks enough detail to decide if it’s an emergency. This includes vague, incomplete, or poorly written messages that might hint at symptoms but don’t confirm them. If a symptom clearly matches the emergency list, classify as Emergency. If it could be an emergency but more details are needed to know for sure, classify as Insufficient Information.

Analysis rules:
- Treat questions as reports of situations that occurred.
- If multiple overlaps and at least one could be an emergency → label Emergency immediately.
- Severity cues (e.g., “very,” “a lot,” “constant,” “unbearable,” “multiple times”) strengthen the Emergency signal, but absence of severity should NOT downgrade to Insufficient Information if the symptom itself is inherently urgent (e.g., bleeding, localised pain, difficulty breathing, high fever, jaundice).
- For ambiguous mentions (e.g., “bleeding” without source), assume worst-case relevant to Knowledge Base (vaginal bleeding).
- For cases where localised pain (e.g., such as headache, back pain, leg pain, etc), bleeding, dizziness is mentioned label it as Emergency.
- For infant feeding issues, if feeding has stopped or the baby is “not feeding,” treat as Emergency unless clearly stated otherwise.
- If information is so vague that no symptoms can be matched at all, then label Insufficient Information.
- Do not penalize informal, chatty, or ungrammatical messages if a symptom match is still possible.
- Use the Knowledge Base as the primary guide for Emergency symptoms, but allow for reasonable clinical interpretation when appropriate.
- Output is routing support, not medical advice; do not give care instructions.

Output JSON only (no extra text, no code fences):
{
  "triage_rules": "<1 –2 sentence triage rules used to decide the label>",
  "label": "Emergency" | "Non-Emergency" | "Insufficient Information",
  "reasoning_summary": "<1 –2 sentence explanation for label in English>"
}

Knowledge Base:
[Emergency symptom list provided here]

\end{lstlisting}

\newpage
\section{Reasoning Trace Analysis: Correct Comprehension, Different Triage}
\label{sec:reasoning-trace-example}

To illustrate how script alone influences triage decisions despite equivalent comprehension, we present a representative case.
 
\begin{table}[H]
\small
\centering
\begin{tabular}{p{0.22\columnwidth} p{0.7\columnwidth}}
\toprule
\textbf{English Translation} & I just had colon surgery three days ago. My stomach is hurting, and I have a fever. What should I do? \\
\midrule
\textbf{Ground Truth} & \texttt{emergent} \\
\bottomrule
\end{tabular}
\end{table}
 
\vspace{0.5em}
 
\begin{table}[H]
\small
\centering
\caption{Hindi Native reasoning trace.}
\begin{tabular}{p{0.22\columnwidth} p{0.7\columnwidth}}
\toprule
\textbf{Script} & Hindi Native \\
\textbf{Predicted Label} & \texttt{emergent} \\
\midrule
\textbf{Reasoning Trace} &
Okay, let's see. The user is saying they had \textbf{colon surgery three days ago} and now they have a \textbf{stomach ache} and a \textbf{fever}. They're asking what they should do.

First, I need to determine the urgency. The key points here are recent surgery and current symptoms. Post-surgery complications can be serious. Pain and fever after surgery might indicate an infection or other issues like a surgical site infection, internal bleeding, or maybe something like a blockage. Since it's only been three days, that's a critical period where complications can arise. The fever is a red flag because it can signify infection. The pain is also concerning because it's not normal after surgery to have increasing pain. 

The triage labels are emergent, conditionally\_emergent, or non\_emergent. Emergent would be if they need immediate ER care. Conditionally emergent if it depends on other factors not mentioned. Non-emergent if it's not urgent.

In this case, the user is showing symptoms that could be serious post-op complications. However, without more details like the severity of the pain, fever temperature, or other symptoms (like vomiting, inability to pass gas, etc.), it's possible but not certain. The presence of both pain and fever after surgery is a common sign of infection, which does require prompt medical attention. But since the message doesn't specify if the fever is high or if there are other symptoms, maybe it's conditionally emergent. Wait, but the guidelines for triage after surgery usually advise contacting a doctor immediately if there's fever and pain. So maybe emergent. 

But the rules say to use only the information present. The user is three days post-op with pain and fever. Those are clear signs that they should seek urgent care. So the correct label would be emergent because they need to go to the ER or call emergency services right away. The reason is that post-surgical complications like infection or internal bleeding can be life-threatening if not treated immediately. So the answer should be emergent. \\
 
\bottomrule
\end{tabular}
\end{table}

\begin{table}[H]
\small
\centering
\caption{Hindi Romanized reasoning trace.}
\begin{tabular}{p{0.22\columnwidth} p{0.7\columnwidth}}
\toprule
\textbf{Script} & Hindi Romanized \\
\textbf{Predicted Label} & \texttt{conditionally\_emergent} \\
\midrule
\textbf{Reasoning Trace} &
Okay, let's see. The user is writing in Hindi, and they mention that they had a \textbf{colon surgery three days ago}. They're experiencing \textbf{stomach pain} and a \textbf{fever}. The question is what they should do. 
First, I need to determine the urgency. The user had recent surgery, and now has pain and fever. Post-surgical complications can be serious. Fever could indicate an infection, which is a common complication after surgery. Stomach pain might be related to the surgery site. These symptoms could be signs of something that needs immediate attention, like an infection or internal bleeding. However, the message doesn't mention other critical symptoms like severe bleeding, difficulty breathing, or loss of consciousness. But since the fever and pain are present after surgery, which are red flags, this could be an emergency. However, the exact cause isn't clear from the message. The user might need to see a doctor urgently, but without more context, like the severity of the fever or pain, it's conditionally emergent. The presence of post-op complications requires immediate care, but the message lacks details to confirm the exact risk level. So the label would be conditionally\_emergent. \\
 
\bottomrule
\end{tabular}
\end{table}

%% file: references.bib
@inproceedings{diekmann2025safetyqa,
  author    = {Diekmann, Yella and Fensore, Chase M. and Carrillo-Larco, Rodrigo M. and Pradhan, Nishant and Appana, Bhavya and Ho, Joyce C.},
  title     = {Evaluating Safety of Large Language Models for Patient-facing Medical Question Answering},
  booktitle = {Proceedings of the 4th Machine Learning for Health Symposium},
  series    = {Proceedings of Machine Learning Research},
  volume    = {259},
  pages     = {267--290},
  year      = {2025},
  publisher = {PMLR},
  url       = {https://proceedings.mlr.press/v259/diekmann25a.html}
}

@article{draelos2025unsafeqa,
  author  = {Draelos, Rachel L. and Afreen, Samina and Blasko, Barbara and Brazile, Tiffany L. and Chase, Natasha and Desai, Dimple Patel and Evert, Jessica and Gardner, Heather L. and Herrmann, Lauren and House, Aswathy Vaikom and Kass, Stephanie and Kavan, Marianne and Khemani, Kirshma and Koire, Amanda and McDonald, Lauren M. and Rabeeah, Zahraa and Shah, Amy},
  title   = {Large language models provide unsafe answers to patient-posed medical questions},
  journal = {arXiv preprint arXiv:2507.18905},
  year    = {2025},
  doi     = {10.48550/arXiv.2507.18905},
  url     = {https://arxiv.org/abs/2507.18905}
}

@article{bedi2024testing,
  author  = {Bedi, Suhana and Liu, Yutong and Orr-Ewing, Lucy and Dash, Dev and Koyejo, Sanmi and Callahan, Alison and Fries, Jason A. and Wornow, Michael and Swaminathan, Akshay and Lehmann, Lisa Soleymani and Hong, Hyo Jung and Kashyap, Mehr and Chaurasia, Akash R. and Shah, Nirav R. and Singh, Karandeep and Tazbaz, Troy and Milstein, Arnold and Pfeffer, Michael A. and Shah, Nigam H.},
  title   = {Testing and Evaluation of Health Care Applications of Large Language Models: A Systematic Review},
  journal = {JAMA},
  year    = {2025},
  volume  = {333},
  number  = {4},
  pages   = {319--328},
  doi     = {10.1001/jama.2024.21700},
  url     = {https://jamanetwork.com/journals/jama/fullarticle/2825147}
}

@article{ren2024primaryconcerns,
  author  = {Ren, Yang and Wu, Yuqi and Fan, Jungwei W. and Khurana, Aditya and Fu, Sunyang and Wu, Dezhi and Liu, Hongfang and Huang, Ming},
  title   = {Automatic Uncovering of Patient Primary Concerns in Portal Messages Using a Fusion Framework of Pretrained Language Models},
  journal = {Journal of the American Medical Informatics Association},
  year    = {2024},
  volume  = {31},
  number  = {8},
  pages   = {1714--1724}
}

@article{chen2025raec,
  author        = {Chen, Wenyuan and Nateghi Haredasht, Fateme and Black, Kameron C. and Grolleau, Francois and Alsentzer, Emily and Chen, Jonathan H. and Ma, Stephen P.},
  title         = {Retrieval-Augmented Guardrails for {AI}-Drafted Patient-Portal Messages: Error Taxonomy Construction and Large-Scale Evaluation},
  journal       = {arXiv preprint arXiv:2509.22565},
  year          = {2025},
  eprint        = {2509.22565},
  archivePrefix = {arXiv},
  primaryClass  = {cs.CL},
  doi           = {10.48550/arXiv.2509.22565},
  url           = {https://arxiv.org/abs/2509.22565}
}

@article{borkowski2023chatgpt,
  author  = {Borkowski, Andrew A. and Jakey, Colleen E. and Mastorides, Stephen M. and Kraus, Ana L. and Vidyarthi, Gitanjali and Viswanadhan, Narayan and Lezama, Jose L.},
  title   = {Applications of ChatGPT and Large Language Models in Medicine and Health Care: Benefits and Pitfalls},
  journal = {Federal Practitioner},
  year    = {2023},
  volume  = {40},
  number  = {6},
  pages   = {170--173},
  doi     = {10.12788/fp.0386},
  url     = {https://pubmed.ncbi.nlm.nih.gov/37860071/}
}

@article{singhal2025medpalm,
  author  = {Singhal, Karan and Tu, Tao and Gottweis, Juraj and Sayres, Rory and Wulczyn, Ellery and Amin, Mohamed and Hou, Le and Clark, Kevin and Pfohl, Stephen R. and Cole-Lewis, Heather and Neal, Darlene and Rashid, Qazi Mamunur and Schaekermann, Mike and Wang, Amy and Dash, Dev and Chen, Jonathan H. and Shah, Nigam H. and Lachgar, Sami and Mansfield, Philip Andrew and Prakash, Sushant and Green, Bradley and Dominowska, Ewa and Ag{\"u}era y Arcas, Blaise and Toma{\v{s}}ev, Nenad and Liu, Yun and Wong, Renee and Semturs, Christopher and Mahdavi, S. Sara and Barral, Joelle K. and Webster, Dale R. and Corrado, Greg S. and Matias, Yossi and Azizi, Shekoofeh and Karthikesalingam, Alan and Natarajan, Vivek},
  title   = {Toward Expert-Level Medical Question Answering with Large Language Models},
  journal = {Nature Medicine},
  year    = {2025},
  volume  = {31},
  url     = {https://www.nature.com/articles/s41591-024-03423-7}
}

@article{kaur2025portal,
  author  = {Kaur, Amarpreet and Budko, Alexander and Liu, Katrina and Eaton, Eric and Steitz, Bryan D. and Johnson, Kevin B.},
  title   = {Automating Responses to Patient Portal Messages Using Generative {AI}},
  journal = {Applied Clinical Informatics},
  year    = {2025},
  volume  = {16},
  number  = {3},
  pages   = {718--731},
  doi     = {10.1055/a-2565-9155},
  url     = {https://pubmed.ncbi.nlm.nih.gov/40132987/}
}

@inproceedings{husain2024romansetu,
  title     = {RomanSetu: Efficiently unlocking multilingual capabilities of Large Language Models via {R}omanization},
  author    = {Husain, Jaavid Aktar and Dabre, Raj and Kumar, Aswanth and Gala, Jay and Jayakumar, Thanmay and Puduppully, Ratish and Kunchukuttan, Anoop},
  booktitle = {Proceedings of the 62nd Annual Meeting of the Association for Computational Linguistics (Volume 1: Long Papers)},
  year      = {2024},
  month     = aug,
  address   = {Bangkok, Thailand},
  publisher = {Association for Computational Linguistics},
  pages     = {15593--15615}
}

@article{masanneck2024triage,
  title   = {Triage Performance Across Large Language Models and ChatGPT},
  author  = {Masanneck, Lucas and others},
  journal = {Journal of Medical Internet Research},
  year    = {2024},
  url     = {https://www.jmir.org/2024/1/e53297/}
}

@article{liu2025detecting,
  title   = {Detecting emergencies in patient portal messages using large language models and knowledge graph-based retrieval-augmented generation},
  author  = {Liu, Siru and others},
  journal = {J Am Med Inform Assoc.},
  year    = {2025},
  url     = {https://pubmed.ncbi.nlm.nih.gov/40220286/}
}

@inproceedings{wang2024dualscript,
  title     = {Enhancing {H}indi Feature Representation through Fusion of Dual-Script Word Embeddings},
  author    = {Wang, Lianxi and Tian, Yujia and Chen, Zhuowei},
  booktitle = {Proceedings of the 2024 Joint International Conference on Computational Linguistics, Language Resources and Evaluation (LREC-COLING 2024)},
  year      = {2024},
  month     = may,
  address   = {Torino, Italia},
  publisher = {ELRA and ICCL},
  pages     = {5966--5976},
  url       = {https://aclanthology.org/2024.lrec-main.528/}
}

@article{mehta2024human,
  title={Human resource shortage in India’s health sector: a scoping review of the current landscape},
  author={Mehta, Vini and Ajmera, Puneeta and Kalra, Sheetal and Miraj, Mohammad and Gallani, Ruchika and Shaik, Riyaz Ahamed and Serhan, Hashem Abu and Sah, Ranjit},
  journal={BMC Public Health},
  volume={24},
  number={1},
  pages={1368},
  year={2024},
  publisher={Springer}
}

@article{medqa,
  title={What disease does this patient have? a large-scale open domain question answering dataset from medical exams},
  author={Jin, Di and Pan, Eileen and Oufattole, Nassim and Weng, Wei-Hung and Fang, Hanyi and Szolovits, Peter},
  journal={Applied Sciences},
  volume={11},
  number={14},
  pages={6421},
  year={2021},
  publisher={MDPI}
}

@InProceedings{MedMCQA,
  title = 	 {MedMCQA: A Large-scale Multi-Subject Multi-Choice Dataset for Medical domain Question Answering},
  author =       {Pal, Ankit and Umapathi, Logesh Kumar and Sankarasubbu, Malaikannan},
  booktitle = 	 {Proceedings of the Conference on Health, Inference, and Learning},
  pages = 	 {248--260},
  year = 	 {2022},
  editor = 	 {Flores, Gerardo and Chen, George H and Pollard, Tom and Ho, Joyce C and Naumann, Tristan},
  volume = 	 {174},
  series = 	 {Proceedings of Machine Learning Research},
  month = 	 {07--08 Apr},
  publisher =    {PMLR}
}

@inproceedings{jin-etal-2019-pubmedqa,
    title = "{P}ub{M}ed{QA}: A Dataset for Biomedical Research Question Answering",
    author = "Jin, Qiao  and
      Dhingra, Bhuwan  and
      Liu, Zhengping  and
      Cohen, William  and
      Lu, Xinghua",
    editor = "Inui, Kentaro  and
      Jiang, Jing  and
      Ng, Vincent  and
      Wan, Xiaojun",
    booktitle = "EMNLP-IJCNLP 2019",
    month = nov,
    year = "2019",
    address = "Hong Kong, China",
    pages = "2567--2577"
}

@inproceedings{verma2025milu,
  title={Milu: A multi-task indic language understanding benchmark},
  author={Verma, Sshubam and Khan, Mohammed Safi Ur Rahman and Kumar, Vishwajeet and Murthy, Rudra and Sen, Jaydeep},
  booktitle={NAACL 2025},
  year={2025}
}

@inproceedings{singh2025indic,
  title={Indic qa benchmark: A multilingual benchmark to evaluate question answering capability of llms for indic languages},
  author={Singh, Abhishek Kumar and Kumar, Vishwajeet and Murthy, Rudra and Sen, Jaydeep and Mittal, Ashish and Ramakrishnan, Ganesh},
  booktitle={Findings of NAACL 2025},
  year={2025}
}

@article{singh2024indicgenbench,
  title={Indicgenbench: A multilingual benchmark to evaluate generation capabilities of llms on indic languages},
  author={Singh, Harman and Gupta, Nitish and Bharadwaj, Shikhar and Tewari, Dinesh and Talukdar, Partha},
  journal={arXiv preprint arXiv:2404.16816},
  year={2024}
}

@inproceedings{kumar2022indicnlg,
  title={IndicNLG benchmark: Multilingual datasets for diverse NLG tasks in Indic languages},
  author={Kumar, Aman and Shrotriya, Himani and Sahu, Prachi and Mishra, Amogh and Dabre, Raj and Puduppully, Ratish and Kunchukuttan, Anoop and Khapra, Mitesh M and Kumar, Pratyush},
  booktitle={EMNLP 2022},
  pages={5363--5394},
  year={2022}
}

@inproceedings{manes-etal-2024-k,
    title = "K-{QA}: A Real-World Medical {Q}{\&}{A} Benchmark",
    author = "Manes, Itay  and
      Ronn, Naama  and
      Cohen, David  and
      Ilan Ber, Ran  and
      Horowitz-Kugler, Zehavi  and
      Stanovsky, Gabriel",
    booktitle = "Proceedings of the 23rd Workshop on Biomedical Natural Language Processing",
    month = aug,
    year = "2024",
    address = "Bangkok, Thailand",
    pages = "277--294"
}

@article{arora2025healthbench,
  title={Healthbench: Evaluating large language models towards improved human health},
  author={Arora, Rahul K and Wei, Jason and Hicks, Rebecca Soskin and Bowman, Preston and Qui{\~n}onero-Candela, Joaquin and Tsimpourlas, Foivos and Sharman, Michael and Shah, Meghan and Vallone, Andrea and Beutel, Alex and others},
  journal={arXiv preprint},
  year={2025}
}

@article{he2020meddialog,
  title={Meddialog: Two large-scale medical dialogue datasets},
  author={He, Xuehai and Chen, Shu and Ju, Zeqian and Dong, Xiangyu and Fang, Hongchao and Wang, Sicheng and Yang, Yue and Zeng, Jiaqi and Zhang, Ruisi and Zhang, Ruoyu and others},
  journal={arXiv preprint arXiv:2004.03329},
  year={2020}
}
